\documentclass{article}

\PassOptionsToPackage{numbers, compress}{natbib}

 \usepackage[main, final]{neurips_2026}

\usepackage[utf8]{inputenc} 
\usepackage[T1]{fontenc}    
\usepackage{url}            
\usepackage{booktabs}       
\usepackage{amsfonts}       
\usepackage{nicefrac}       
\usepackage{microtype}      
\usepackage{xcolor}         
\usepackage{pifont}

\usepackage{graphicx}
\usepackage{multirow, multicol}
\usepackage{mathtools}
\usepackage{amsmath,amsfonts,amssymb,amsthm,bbm}
\usepackage{mathabx}
\usepackage{algorithm}
\usepackage{algorithmic}
\usepackage{colortbl}
\usepackage{amssymb}
\definecolor{mygray}{gray}{0.9}
\usepackage{wrapfig} 
\usepackage{caption}
\usepackage{subcaption}
\usepackage{enumitem}
\usepackage[table]{xcolor}
\usepackage{hyperref}       

\newcommand{\vllm}{{Video-LLMs}}
\newcommand{\method}{{EvoGround}}

\newcommand{\cmark}{\ding{51}}%
\newcommand{\xmark}{\ding{55}}%
\newcommand{\eg}{\textit{e.g.}}
\newcommand{\ie}{\textit{i.e.}}

\title{EvoGround: Self-Evolving Video Agents \\ for Video Temporal Grounding}

\author{
  \bf Minjoon Jung\textsuperscript{\rm 1,2}~~~
    Byoung-Tak Zhang\textsuperscript{\rm 2}~~~ 
    Lorenzo Torresani\textsuperscript{\rm 1} \vspace{1.5mm} \\
   \textsuperscript{1}Northeastern University~~~
   \textsuperscript{2}Seoul National University \vspace{1.5mm}
   \\ \tt\small{\{jung.minjo,\,l.torresani\}@northeastern.edu}~~~\tt\small{\{btzhang\}@bi.snu.ac.kr}
}

\begin{document}
\maketitle

\begin{abstract}
Video temporal grounding (VTG) takes an untrimmed video and a natural-language query as input and localizes the temporal moment that best matches the query. Existing methods rely on large, task-specific datasets requiring costly manual annotation. We introduce EvoGround, a framework of two coupled self-evolving agents, a \textit{proposer} and a \textit{solver}, that learn temporal grounding from raw videos without any task-specific annotations. The proposer generates query--moment pairs from raw videos, while the solver learns to ground them and feeds back signals that improve the proposer in return. Through this self-reinforcing reinforcement-learning loop, the two agents are initialized from the \emph{same} backbone and mutually improve across iterations. Trained on 2.5K unlabeled videos, EvoGround matches or surpasses most fully supervised models across five VTG benchmarks, while emerging as a state-of-the-art fine-grained video captioner. We publicly release all resources at \url{https://minjoong507.github.io/projects/EvoGround/}.
\end{abstract}    
\section{Introduction}
Accurate temporal grounding underpins a wide range of applications, including video retrieval~\cite{jung2022modal}, highlight detection~\cite{lei2021qv}, and grounded question answering~\cite{xiao2024can}, and is increasingly relied upon as a core module in complex video understanding pipelines~\cite{shangguan2024tomato, jung2026egoexo}. Given an untrimmed video and a natural language query, video temporal grounding (VTG)~\cite{gao2017tall} aims to localize the temporal moment within the video that corresponds to the query. Despite remarkable progress, existing models typically rely on large-scale human-labeled data for training, fundamentally limiting their scalability and applicability.

Collecting labeled data for VTG is particularly burdensome. Annotators must manually identify temporal segments and provide corresponding natural language descriptions within untrimmed videos, a time-consuming process that inherently limits scalability. This often leads to a sparsity issue~\cite{zeng2020dense, li2024exploiting}, where only a fraction of video events are covered, amplifying the narrowness of the training distribution. Beyond scalability, obtaining high-quality annotations remains non-trivial: queries are often ambiguous or misaligned with the underlying visual content, causing models to learn spurious correlations or exploit shortcuts rather than being grounded in true visual evidence~\cite{yuan2021closer, otani2020uncovering, jung2025consistency}.

\emph{Can a model reliably learn temporal information in videos without manual labels?} At first glance, this seems unlikely, as temporal grounding requires precise, fine-grained reasoning that one might assume can only emerge from explicit supervision. We challenge this assumption with \method, a framework of two coupled self-evolving agents that learn to ground temporal moments through mutual interaction without any task-specific video annotation. 

\begin{figure}[t]
    \centering
    \includegraphics[width=0.99\linewidth]{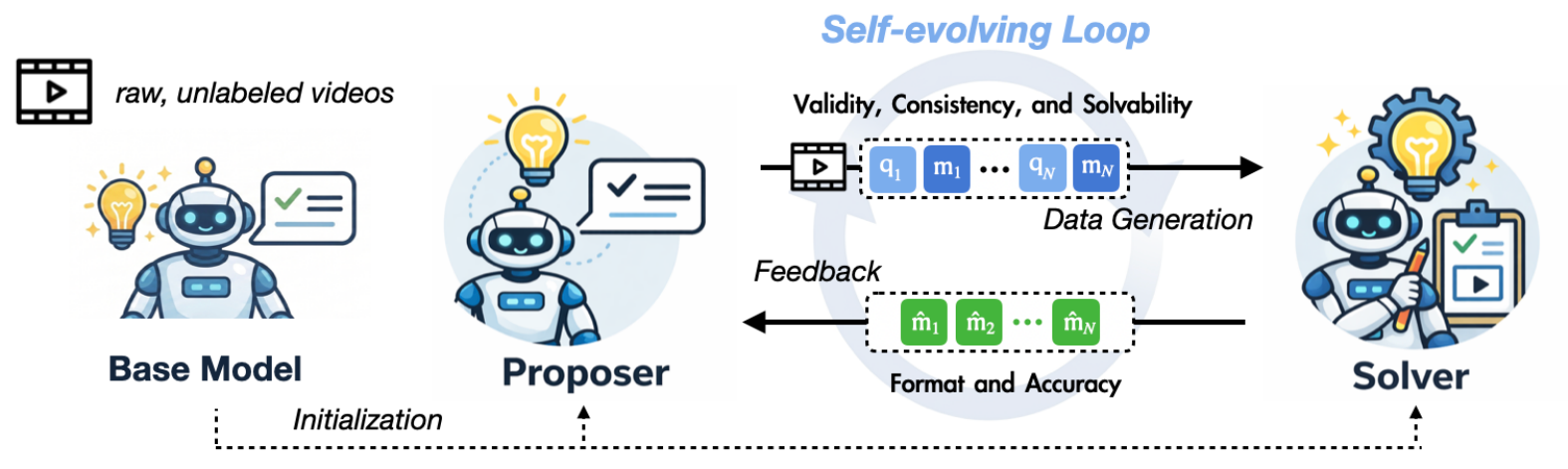}
    \caption{
    \textbf{\method: a self-evolving loop, with unlabeled videos.} A proposer and a solver, both initialized from the same base model, co-evolve through reinforcement learning. The proposer generates query ($q$)--moment ($m$) pairs from a raw video; the solver grounds them and produces predictions ($\hat{m}$) that feed back as a learning signal. Dedicated reward designs guide each agent.
    }
    \label{fig: teaser}
    \vspace{-3mm}
\end{figure}

As shown in Figure~\ref{fig: teaser}, \method~consists of a \textit{proposer} and a \textit{solver}, evolved through an iterative two-stage process via reinforcement learning~(RL), with both agents initialized from the same base model. Rather than relying on human annotators, the proposer takes on their role: given a raw video, it identifies temporal moments and generates corresponding natural language queries. The solver then attempts to ground these queries back to their moments, and its success or failure is fed back to the proposer as a learning signal.  

In the first stage, we encourage the proposer to provide valid, semantically consistent, and solvable query-moment pairs; in the second stage, the solver is trained on the data generated by the updated proposer. Through this self-reinforcing loop with dedicated reward designs for each agent, they challenge and sharpen each other: a better proposer generates more meaningful and solvable queries, which in turn train a better solver, whose improved feedback further raises the bar for the proposer. This virtuous cycle, driven entirely by reinforcement learning, improves over successive iterations before saturating.

Despite never being exposed to any single manual grounding annotation during the self-evolution phases, \method~matches or outperforms most fully-supervised models across five VTG benchmarks: Charades-STA~\cite{gao2017tall}, ActivityNet-Captions~\cite{krishna2017dense}, TVGBench~\cite{wang2025timer1}, ReXTime~\cite{chen2024rextime}, and E.T.Bench~\cite{liu2024etbench}. Moreover, \method~demonstrates its strong fine-grained video captioning capability on TemporalBench~\cite{cai2024temporalbench}, suggesting that the self-evolution loop fosters a broader and more grounded understanding of video content.

Our primary contributions are summarized as follows:
\begin{itemize}[leftmargin=10pt]
    \item We propose \method, a framework of self-evolving agents that learns temporal information from raw videos without any task-specific annotation.
    \item \method~adopts a proposer--solver loop in which the two agents, initialized from the same backbone, iteratively challenge and improve each other.
    \item \method~achieves accuracy on par with, and in five VTG benchmarks surpassing, most fully supervised models, while demonstrating its fine-grained video captioning on TemporalBench.
\end{itemize}
\section{Related Work}
\noindent \textbf{Video Temporal Grounding (VTG).} VTG is a task that aims to localize temporal moments from given natural language sentences within untrimmed videos. VTG models typically encode multimodal features using a pre-trained encoder (\eg, CLIP~\cite{radford2021learning}) and adopt proposal-based~\cite{zhang20202dtan}, proposal-free~\cite{zhang2020vslnet, mun2020local}, and DETR-based methods~\cite{lei2021qv, Jung_2025_WACV} for prediction. However, these models are inherently less generalizable and limited by the capacity of pre-trained encoders. Such limitations have motivated a shift toward more general and scalable approaches that employ large language models (LLMs).

\noindent \textbf{Reducing Annotation Cost in VTG.} Prior work reduces but does not remove annotation cost. They typically synthesize pseudo query--moment pairs through fixed pipelines, leveraging pre-trained encoders~\cite{nam2021zero, jung2022modal}, prompted LLMs~\cite{deng2025btdp}, or post-hoc refinement of generated annotations~\cite{bao2025vidgroup}. Because the generator is never updated, labels can be corrected but the proposal distribution cannot improve. \method~optimizes the generator itself, so data quality improves across iterations (Section~\ref{sec:analysis}, Appendix Table~\ref{appendix:tbl:zeroshot_prior}).

\noindent \textbf{Video Large Language Models (Video-LLMs).} Video-LLMs leverage the powerful capabilities of LLMs~\cite{chiang2023vicuna, touvron2023llama} to handle a wide range of downstream tasks, from question answering~\cite{Maaz2023VideoChatGPT, li2023videochat, song2024moviechat} to fine-grained temporal understanding~\cite{vtimellm, guo2024vtgllm, timechat, jung2025consistency, jung2026egoexo, zeng2024timesuite}. Recently, inspired by the success of RL (\eg, DeepSeek-R1~\cite{guo2025deepseek}), RL-based Video-LLMs~\cite{feng2025videor1, wang2025timer1, li2025videochatr1, yan2025videochatr_1_5} have emerged and demonstrate superior spatio-temporal reasoning. However, existing models typically rely on labeled data and often require strong teacher models for guidance. For instance, VideoChat-R1.5~\cite{yan2025videochatr_1_5} leverages DeepSeek to ensure training data quality, and VideoChat-R1~\cite{li2025videochatr1} utilizes Qwen2.5-72B~\cite{qwen2024qwen2} to encourage alignment between model outputs and caption supervision. In contrast, \method~avoids such reliance and learns temporal information directly from raw videos through iterative self-reinforcing loops.

\noindent \textbf{Self-Evolving Agents.} 
Self-evolving frameworks~\cite{yue2026drzero, huang2025rzero, zhao2025absolute} have improved LLMs with minimal or no training data. Recent works~\cite{li2026mmzero, he2025visplay, thawakar2025evolmm} have expanded this mechanism from text to vision-language. We adopt this loop, and extend it to video temporal understanding. Unlike text-domain self-play~\cite{huang2025rzero}, video requires rewards that verify each task against the input and grade it by overlap rather than correctness. Both agents share the same backbone, with no stronger video grounder or teacher in the loop. Any gain reflects the procedure itself rather than distillation.
\section{Method}
\begin{figure*}[t]
        \centering
        \includegraphics[width=1.0\linewidth]{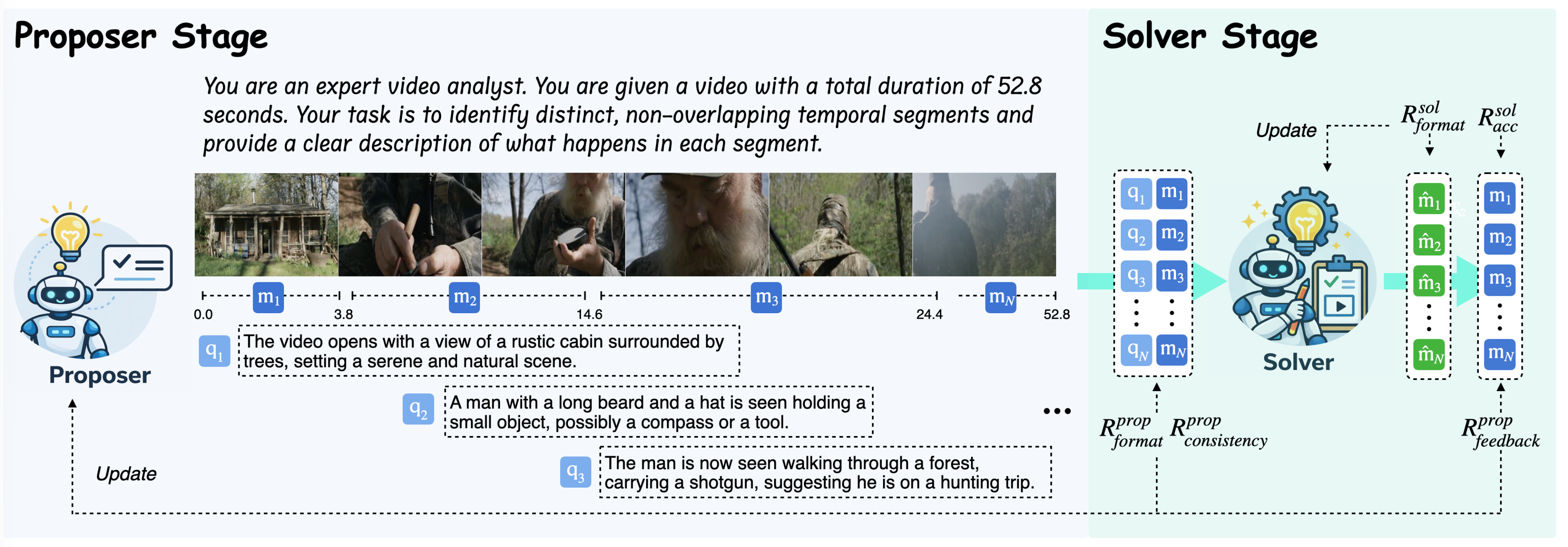}
    \caption{
    \textbf{Overview of \method.} Both agents start from the same backbone. The proposer is updated via three rewards: $R^{\text{prop}}_{\text{format}}$ (\emph{validity}), $R^{\text{prop}}_{\text{consistency}}$ (\emph{consistency}, computed with SigLIP-2), and $R^{\text{prop}}_{\text{feedback}}$ (\emph{solvability}, derived from the solver's tIoU). The solver is updated via $R^{\text{sol}}_{\text{format}}$ and $R^{\text{sol}}_{\text{acc}}$. Stages alternate: the proposer's pairs train the solver, the solver's predictions sharpen the proposer.
    }
    \label{fig: method}
    \vspace{-3mm}
\end{figure*}
As shown in Figure~\ref{fig: method}, we introduce \method, consisting of two agents: a \textit{proposer} and a \textit{solver}. The proposer identifies candidate temporal events from raw videos and generates corresponding query-moment pairs, while the solver learns to ground temporal moments using the generated data. Both agents are initialized from the same backbone~\cite{bai2025qwen2}, and evolve solely through the self-reinforcing loop without any labeled data. Appendix Figure~\ref{appendix:fig:prompt} provides full prompts for each agent.

\subsection{Proposer}
To guide the proposer toward generating high-quality data, we define three key criteria: \emph{validity} (format reward), \emph{consistency} (consistency reward), and \emph{solvability} (feedback reward). (1) Validity: Temporal moments must be defined by numeric timestamps within a valid range (\ie, non-negative and within the video duration) and avoid extremely short intervals or trivial spans that cover the entire video. (2) Consistency: A query--moment pair should be semantically coherent, \ie, the query should accurately describe the visual content of the moment and be discriminative with respect to other candidates, independent of the solver. (3) Solvability: The generated pairs must be learnable by the solver, \ie, the solver should be able to localize the moment from the query. A pair may be semantically consistent yet unsolvable if its temporal boundaries are too ambiguous or imprecise. We design the corresponding reward functions based on these criteria.

For each video, we prompt the proposer to generate a set of temporal moments and corresponding queries. Let the proposer generate $N$ query-moment pairs $\{(q_n, m_n)\}_{n=1}^{N}$, where $q_n$ denotes the $n$-th query and $m_n=(s_n,e_n)$ represents the corresponding temporal moment with start time $s_n$ and end time $e_n$ in seconds. For each query $q_n$, the solver produces a moment prediction $\hat{m}_n=(\hat{s}_n,\hat{e}_n)$. Based on these generated pairs and predictions, we denote the $i$-th generated sample as $\mathbf{s}_i = \{q_n, m_n, \hat{m}_n\}_{n=1}^N$, which serves as the input to all reward functions defined below.

\noindent \textbf{Format Reward.} The format reward verifies whether the proposer generates valid query--moment pairs. To facilitate reliable parsing and evaluation, we prompt the model to provide multiple moments and queries, each enclosed within predefined tags, \texttt{<time>...</time>} and \texttt{<description>...</description>}, respectively. The format reward is defined as:
\begin{equation}
    \label{eq:format_reward}
    R_{\text{format}}^{\text{prop}}(\mathbf{s}_i) = 
    \begin{cases}
    \frac{1}{N}\sum_{n=1}^{N} \mathbb{I}\big( (s_n, e_n) \in \mathcal{V} \big), & \text{if the proposer follows the template}, \\
    0, & \text{otherwise,}
    \end{cases}
\end{equation}
where $\mathbb{I}(\cdot)$ is the indicator function and $\mathcal{V}$ denotes the set of valid temporal moments satisfying: (1) $s_n \geq 0$ and $e_n \leq T_{\text{video}}$ and (2) $e_n - s_n > \tau_{\text{length}}$, where $T_{\text{video}}$ is the video duration and $\tau_{\text{length}}$ specifies the minimum segment length. 
In later iterations, we further condition the format reward on the solver's feedback to ensure that generated query--moment pairs are not only structurally valid but also solvable. We measure the feedback using timestamp-aware IoU (tIoU)~\cite{wang2025timer1}:
\begin{equation}
    \text{tIoU}(m_n, \hat{m}_n) = \mathrm{IoU}(m_n, \hat{m}_n) \cdot \left(1 - \frac{|s_n - \hat{s}_n|}{T_{\text{video}}}\right) \cdot \left(1 - \frac{|e_n - \hat{e}_n|}{T_{\text{video}}}\right),
\end{equation}
which extends standard IoU~\cite{yuan2021closer} by incorporating penalties for deviations in start and end timestamps relative to the video duration, leading to improved boundary sensitivity.
Here, $\mathrm{IoU}$ is defined as:
\begin{equation}
    \text{IoU}(m_n, \hat{m}_n) = \frac{|[s_n, e_n] \cap [\hat{s}_n, \hat{e}_n]|}{|[s_n, e_n] \cup [\hat{s}_n, \hat{e}_n]|}.
\end{equation}
Based on this, we define the \emph{conditioned format reward} as:
\begin{equation}
    \mathbb{I}\big( (s_n, e_n) \in \mathcal{V} \ \land \ \mathrm{tIoU}(m_n, \hat{m}_n) \geq \delta \big), 
    \label{eq:conditioned_format}
\end{equation}
where $\delta \in [0, 1]$ is a threshold that conditions the validity of query–moment pairs on the current solver’s accuracy. This design ensures that the proposer generates pairs that are both structurally valid and learnable at each training step. The format reward in Equation~\ref{eq:format_reward} corresponds to the case of $\delta = 0$.

\noindent \textbf{Consistency Reward.}
The consistency reward measures two complementary properties of each 
query-moment pair: \emph{intra-consistency}, which captures how coherently the query aligns with the frames within its corresponding moment, and \emph{inter-consistency}, which captures how discriminatively the query matches its own moment relative to other candidates. We use SigLIP-2~\cite{tschannen2025siglip}, an image--text encoder carrying no temporal grounding supervision, to compute frame-level similarities $S_{n,t} = \operatorname{cos}(q_n, f_t)$ by uniformly sampling $|m_n|$ frames from each moment at a rate proportional to its duration, and obtain their average $\mu_n$. The intra-consistency score penalizes high standard deviation in $S_{n,t}$, rewarding pairs where the query aligns uniformly with frames in $m_n$:
\begin{equation}
\label{eq:intra-consistency}
\text{Intra}_n = \exp(-\gamma \cdot \sigma_n), \quad 
\sigma_n = \sqrt{\frac{1}{|m_n|} \sum_{t=1}^{|m_n|} (S_{n,t} - \mu_n)^2},
\end{equation}
where $\gamma$ controls sensitivity to alignment fluctuations. However, a pair with uniformly low similarity would still score highly under $\text{Intra}_n$ alone. The inter-consistency score addresses this by measuring how discriminatively the query matches its own moment relative to others, computed as a softmax over cross-moment similarities with temperature $\tau$:
\begin{equation}
    \label{eq:inter-consistency}
    \text{Inter}_n = \frac{\exp(\mu_n / \tau)}{\sum_{j=1}^{N}\exp(\mu_{n,j} / \tau)},
\end{equation}
where $\mu_{n,j}$ denotes the mean similarity between query $q_n$ and moment $m_j$. The final consistency reward is their product, requiring both temporal coherence within the moment and semantic discriminability across moments to be simultaneously satisfied:
\begin{equation}
    \label{eq:consistency-reward}
    R^{\text{prop}}_{\text{consistency}}(\mathbf{s}_i) = \frac{1}{N} 
    \sum_{n=1}^{N} \text{Intra}_n \cdot \text{Inter}_n.
\end{equation}
In Appendix~\ref{appendix:subsec:reward designs}, we provide more details on the intra-consistency formulation and the computation of $|m_n|$. We further explore variants of inter-consistency under different formulations (\eg, hard ranking-based) and find that the softmax formulation consistently outperforms the alternatives.

\noindent \textbf{Feedback Reward.} The feedback reward confirms whether the pair is solvable and is measured by the solver's accuracy:
\begin{equation}
    R_{\text{feedback}}^{\text{prop}}(\mathbf{s}_i) = \frac{1}{N}\sum_{n=1}^{N}\mathrm{tIoU}(m_n, \hat{m}_n),
\end{equation}
where $\hat{m}_n$ denotes the moment predictions from query $q_n$.

\subsection{Solver}
From the query-moment pairs generated by the proposer, each pair is processed independently. We consider a single query--moment pair ($q$, $m$). Given the query $q$, the solver generates multiple rollout samples $\{o_k\}_{k=1}^K$, from which we extract the corresponding predicted moments $\{\hat{m}_k\}_{k=1}^K$.

\noindent \textbf{Format Reward.} To facilitate easy answer extraction, we encourage the solver to follow the pre-defined template: \texttt{<think>...</think>} and \texttt{<answer>...</answer>}, where the \texttt{<think>} block contains the model's intermediate reasoning steps and the \texttt{<answer>} block contains the predicted temporal moment as start and end timestamps. Based on this, the format reward is defined as:
\begin{equation}
    R_{\text{format}}^{\text{sol}}(o_k) =
    \begin{cases}
        1, & \text{if the solver follows the template}, \\
        0, & \text{otherwise}.
    \end{cases}
\end{equation}

\noindent \textbf{Accuracy Reward.} To improve accuracy, we extract the prediction $\hat{m}_k$ from the content within the answer tags and measure accuracy by comparing it with the ground-truth moment $m$:
\begin{equation}
    R_{\text{acc}}^{\text{sol}}(o_k) = \mathrm{tIoU}(m, \hat{m}_k).
\end{equation}
Note that while $R_{\text{acc}}^{\text{sol}}$ and $R_{\text{feedback}}^{\text{prop}}$ share the same formulation, they serve distinct roles: the former optimizes the solver toward accurate grounding, while the latter signals to the proposer whether its generated pairs are learnable.
\begin{figure}[t]
        \centering
        \includegraphics[width=0.99\linewidth]{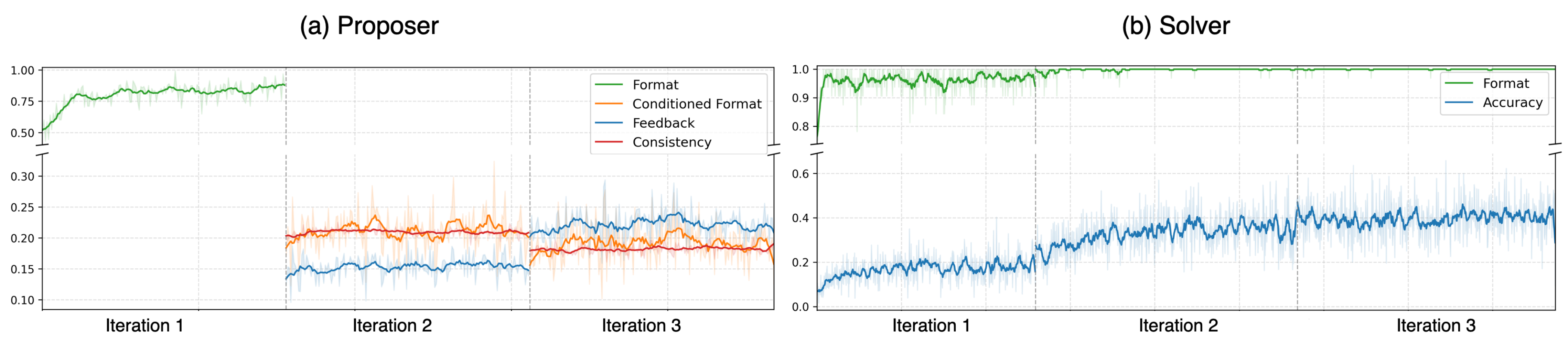}
    \caption{
    \textbf{Reward dynamics across iterations.} We visualize the evolution of the proposer and solver in (a) and (b), respectively. 
    As the proposer evolves over iterations, the solver correspondingly 
    demonstrates progressively higher accuracy. 
    }
    \label{fig:rewards}
    \vspace{-3mm}
\end{figure}

\subsection{Optimization}
\label{sec: optimization}
In this section, we describe how \method~is optimized. \method~alternates between two stages: in the \textit{proposer stage}, the proposer is optimized while the solver is frozen; in the \textit{solver stage}, the solver is updated using query--moment pairs generated by the frozen proposer.

\noindent \textbf{Objective.} We adopt GDPO~\cite{liu2026gdpo} as our RL optimizer, which improves upon GRPO~\cite{guo2025deepseek} by weighting and normalizing each reward component individually before aggregation, rather than summing first and normalizing after. This improves the resolution of the training signals and prevents easier rewards from dominating the optimization, which is critical in our multi-reward setting. We provide the detailed formulation of each objective in Appendix~\ref{appendix:subsec:gdpo}.

\noindent\textbf{Curriculum Design.} 
In the first iteration, we warm up the proposer with the format reward only and update the solver, since neither agent is yet tailored to the task. In subsequent iterations, we apply consistency and feedback rewards to the proposer and adopt GDPO for this multi-reward setting, assigning higher weights to the feedback reward than to the format reward to emphasize more challenging objectives. We also progressively increase $\delta$ (in Equation~\ref{eq:conditioned_format}) across iterations, shifting the focus from coarse matches toward more precise temporal alignments. We validate the effectiveness of these design choices in Section~\ref{sec:analysis}. For the solver, we keep the design simple, applying naive GRPO without additional reward designs or the multi-reward weighting scheme (\ie, GDPO).

\noindent\textbf{Reward Dynamics.} 
Figure~\ref{fig:rewards} visualizes the reward dynamics across iterations. Both agents quickly saturate the format reward in the first iteration. In subsequent iterations the conditioned format reward drops as $\delta$ increases, while the feedback reward steadily improves, tracking the solver's increasing accuracy. This confirms that our curriculum design effectively guides optimization.
\section{Experiments}
\begin{table*}[t]
    \centering
    \caption{
    \textbf{Results on Charades-STA and ActivityNet-Captions.} \method~achieves competitive or superior performance to existing models without using any manual labels, outperforming all SFT-based models and matching the best RL-based models. * denotes our reproduced results.
    }
    \resizebox{\linewidth}{!}{
    \begin{tabular}{l cccc cccc}
    \toprule
    \multirow{2}{*}{\textbf{Method}} 
    & \multicolumn{4}{c}{\textbf{Charades-STA}} 
    & \multicolumn{4}{c}{\textbf{ActivityNet-Captions}} \\
    \cmidrule(lr){2-5} \cmidrule(lr){6-9}
     & R1@0.3 & R1@0.5 & R1@0.7 & mIoU 
     & R1@0.3 & R1@0.5 & R1@0.7 & mIoU \\ 
    \midrule
    
    \textcolor{gray}{\textit{SFT-based Models}} \\
    TimeChat~\cite{timechat} 
    & - & 32.2 & 13.4 & 32.2 
    & 36.2 & 20.2 & 9.5 & 21.8 \\
    
    HawkEye~\cite{wang2024hawkeye} 
    & 50.6 & 31.4 & 14.5 & -
    & 49.1 & 29.3 & 10.7 & - \\
    
    VTimeLLM~\cite{vtimellm} 
    & 51.0 & 27.5 & 11.4 & 31.2 
    & 44.0 & 27.8 & 14.3 & 30.4 \\


    Qwen2.5-VL*~\cite{bai2025qwen2} 
    & 68.5 & 48.8 & 22.5 & 45.0
    & 38.7 & 25.6 & 14.9 & 28.6 \\
    
    
    TimeSuite~\cite{zeng2024timesuite} 
    & 69.9 & 48.7 & 24.0 & - 
    & - & - & - & - \\


    TRACE~\cite{guo2024trace} 
    & 58.6 & 40.3 & 19.4 & 38.7 
    & - & - & - & -
    \\

    ED-VTG~\cite{pramanick2025enrich} 
    & 59.5 & 39.3 & 19.8 & 40.2 
    & 52.1 & 33.1 & 16.0 & 35.2 
    \\
    
    \midrule \textcolor{gray}{\textit{RL-based Models}} \\
    VideoChat-R1~\cite{li2025videochatr1} 
    & - & - & - & - 
    & 50.4 & 32.2 & 16.2 & 34.3 \\
    
    VideoChat-R1.5~\cite{yan2025videochatr_1_5} 
    & - & - & - & - 
    & 52.4 & 32.3 & 16.8 & 35.5 \\
    
    Time-R1~\cite{wang2025timer1} 
    & \bf 78.1 & \bf 60.8 & \underline{35.3} & \bf 58.1 
    & \underline{58.6} & \underline{39.0} & \underline{21.4} & \underline{40.5} \\         
    \midrule
    \bf \method~(Ours) 
    & \underline{77.2} {\footnotesize$\pm$1.3} & \underline{60.5} {\footnotesize$\pm$1.6} & \textbf{35.5} {\footnotesize$\pm$1.5} & \underline{53.1} {\footnotesize$\pm$0.9}
    & \textbf{62.9} {\footnotesize$\pm$0.8} & \textbf{43.6} {\footnotesize$\pm$0.7} & \textbf{25.0} {\footnotesize$\pm$0.6} & \textbf{43.9} {\footnotesize$\pm$0.5} \\
    \bottomrule
    \end{tabular}
    }
    \label{tbl:charades&activity}
\end{table*}
\subsection{Experiment Setup}
\noindent \textbf{Baselines.} We consider \vllm~as baselines, most of which focus on VTG. We categorize them into two folds: SFT-based models, including TimeChat~\cite{timechat}, HawkEye~\cite{wang2024hawkeye}, VTimeLLM~\cite{vtimellm}, Qwen2.5-VL~\cite{bai2025qwen2}, TimeSuite~\cite{zeng2024timesuite}, VideoChat-Flash~\cite{li2024videochatflash}, LITA~\cite{huang2024lita}, TRACE~\cite{guo2024trace}, and ED-VTG~\cite{pramanick2025enrich}, and RL-based models, including VideoChat-R1~\cite{li2025videochatr1}, VideoChat-R1.5~\cite{yan2025videochatr_1_5}, and Time-R1~\cite{wang2025timer1}. For captioning, we include various \vllm, including Qwen2-VL~\cite{qwen2024qwen2}, LLaVA-OneVision~\cite{li2024llavaov}, LLaVA-NeXT-Video~\cite{liu2024llavanext}, InternLM-XC2.5~\cite{zhang2024internlm}, VideoLLaVA~\cite{lin2023videollava}, MiniCPM-V2.6~\cite{yao2024minicpm}, Phi-3.5.Vision~\cite{abdin2024phi}, MA-LLM~\cite{he2024mallm}, and Matryoshka Multimodal Models ($\text{M}^3$)~\cite{cai2024m3}. Models are 7B-scale unless otherwise specified. Among all baselines, \method~is the only model explicitly trained for VTG without task-specific annotations.

\noindent \textbf{Benchmarks.} 
We conduct experiments on five VTG benchmarks: Charades-STA~\cite{gao2017tall}, ActivityNet-Captions~\cite{krishna2017dense}, TVGBench~\cite{wang2025timer1}, ReXTime~\cite{chen2024rextime}, and E.T.Bench~\cite{liu2024etbench}. 
Additionally, we use TemporalBench~\cite{cai2024temporalbench} to evaluate the model's captioning capability. We follow the cross-dataset zero-shot protocol of prior RL-based work~\cite{wang2025timer1}, where no model is fine-tuned on the evaluated benchmarks.

\noindent \textbf{Evaluation Metrics.} For grounding, we use R@$n$, IoU=$m$ metric, which measures the percentage of top-$n$ predicted moments whose IoU with the ground-truth moment exceeds a threshold $m \in \{0.3, 0.5, 0.7\}$, and also report the mean IoU (mIoU) and F1 score. For further reliable comparison, we report 95\% confidence intervals estimated via nonparametric bootstrap (\ie, 1,000 resamples with replacement of the test set\footnote{We exclude ReXTime for this since its ground-truth data is released in the official test server}). Point estimates are computed on the full test set, and intervals correspond to the 2.5th–97.5th percentiles of the bootstrap distribution. For captioning, we adopt classical captioning metrics: CIDEr~\cite{vedantam2015cider}, BLEU~\cite{papineni2002bleu} at various n-gram levels, ROUGE~\cite{lin2004rouge}, and sentence similarity between generated and ground-truth captions with a pre-trained Sentence Transformer~\cite{reimers2019sentencetransformer}.

\noindent \textbf{Implementation Details.} We use Qwen2.5-VL-7B~\cite{bai2025qwen2} as a backbone for a fair comparison with prior RL-based models~\cite{li2025videochatr1, yan2025videochatr_1_5, wang2025timer1} and utilize 2.5K raw videos from TimeRFT~\cite{wang2025timer1}. We perform three iterations to train both agents, and each iteration generates $\sim$9K query--moment pairs, completing in $\sim$24 hours on 4$\times$A100-80GB. For the proposer, we progressively increase $\delta$ ($0 \rightarrow 0.3 \rightarrow 0.5$) and assign reward weights of $w_{\text{format}}=0.5$, $w_{\text{consistency}}=0.5$, and $w_{\text{feedback}}=1.0$. Following~\cite{wang2025timer1}, video frames are sampled at 2 FPS and adaptively resized to 2.8M pixels to control efficiency during both training and testing; results marked with * in the following tables are obtained from the official checkpoint under the same setup. Please refer to Appendix~\ref{appendix:subsec:additional implementation details} for more implementation details.
\begin{table*}[t]
    \centering
    \caption{
    \textbf{Results on TVGBench, ReXTime, and E.T.Bench.} Across diverse benchmarks spanning varying video durations and domains, \method~surpasses SFT-based models and remains competitive with RL-based models trained on human-annotated data. * denotes our reproduced results.
    }
    \resizebox{\linewidth}{!}{
    \begin{tabular}{l cccc ccc c}
    \toprule
    \multirow{2}{*}{\textbf{Method}} & \multicolumn{4}{c}{\textbf{TVGBench}} & \multicolumn{3}{c}{\textbf{ReXTime}} & \textbf{E.T.Bench} \\ 
    \cmidrule(lr){2-5}
    \cmidrule(lr){6-8}
    \cmidrule(lr){9-9}
     & R1@0.3 & R1@0.5 & R1@0.7 & mIoU & R1@0.3 & R1@0.5 & mIoU & $\text{TVG}_{\textit{F1}}$ \\ 
    \midrule
    
    \textcolor{gray}{\textit{SFT-based Models}} \\ 
    
    TimeChat~\cite{timechat} & 22.4 & 11.9 & 5.3 & - & 14.4 & 7.6 & 11.6 & 26.2 \\

    VTimeLLM~\cite{vtimellm} & - & - & - & - & 28.8 & 17.4 & 20.1 & 7.6 \\
    
    Qwen2.5-VL*~\cite{bai2025qwen2} 
    & 28.1 & 19.5 & 10.5 & 20.4 
    & 16.3 & 11.2 & 13.3 
    & 46.6 \\
    
    LITA-13B~\cite{huang2024lita}
    & - & - & - & - 
    & 29.4 & 16.2 & 21.4 
    & 22.2 \\

    TimeSuite~\cite{zeng2024timesuite} 
    & 31.1 & 18.0 & 8.9 & -
    & - & - & - 
    & - \\

    VideoChat-Flash~\cite{li2024videochatflash}
    & 32.8 & 19.8 & 10.4 & -
    & - & - & - 
    & - \\
    
    TRACE~\cite{guo2024trace} 
    & 37.0 &25.5 & 14.6  & - 
    & - & - & - 
    & - \\
    \midrule \textcolor{gray}{\textit{RL-based Models}} \\

    VideoChat-R1.5*~\cite{yan2025videochatr_1_5} 
    & 31.5 & 20.8 & 11.3 & 22.0 
    & 28.3 & 18.0 & 20.5 
    & 50.3
    \\
    
    Time-R1*~\cite{wang2025timer1} 
    & \underline{39.3} & \underline{28.0} & \textbf{16.0} & \underline{27.8} 
    & \underline{32.2} & \underline{22.1} & \underline{24.1}
    & \textbf{69.4} \\

    
    \midrule
    
    \bf \method~(Ours) 
    & \textbf{42.1} {\footnotesize$\pm$3.3} & \textbf{28.5} {\footnotesize$\pm$3.1} & \underline{15.3} {\footnotesize$\pm$2.6} & \textbf{29.4} {\footnotesize$\pm$2.1}
    & \bf{33.5} & \bf{22.4} & \bf{25.5} 
    & \underline{69.0} {\footnotesize$\pm$2.2}
    \\
    
    \bottomrule
    \end{tabular}
    }
    \label{tbl:tvgbench&rextime&etbench}
\end{table*}
\begin{table*}[t]
\centering
\caption{
    \textbf{Results on TemporalBench.} \method~outperforms all prior models across all captioning metrics, demonstrating broader video understanding beyond temporal grounding.
}
\resizebox{\textwidth}{!}{
\begin{tabular}{l ccccccc}
    \toprule
    \multirow{2}{*}{\bf Method} & \multicolumn{7}{c}{\bf TemporalBench} \\
    \cmidrule(lr){2-8}
    & Similarity & CIDEr & ROUGE & BLEU\_1 & BLEU\_2 & BLEU\_3 & BLEU\_4 \\
    \midrule
    Qwen2-VL~\cite{qwen2024qwen2} & 51.9 & 6.9 & 18.0 & 12.5 & 6.1 & 3.0 & 1.6 \\
    LLaVA-OneVision~\cite{li2024llavaov} & 50.1 & 0.3 & 14.5 & 11.1 & 5.1 & 2.2 & 1.1 \\
    LLaVA-NeXT-Video~\cite{liu2024llavanext} & 50.1 & 2.3 & 15.8 & 18.1 & 7.0 & 2.6 & 1.1 \\
    InternLM-XC2.5~\cite{zhang2024internlm} & 52.4 & 2.3 & 15.9 & 17.8 & 7.1 & 2.8 & 1.2 \\
    VideoLLaVA~\cite{lin2023videollava} & 46.0 & 4.5 & 16.9 & 12.6 & 5.4 & 2.3 & 1.0 \\
    MiniCPM-V2.6~\cite{yao2024minicpm} & 47.2 & 1.5 & 14.2 & 15.5 & 5.4 & 1.9 & 0.8 \\
    Phi-3.5-Vision~\cite{abdin2024phi} & 42.9 & 3.7 & 16.5 & 20.4 & 8.4 & 3.4 & 1.6 \\
    MA-LMM~\cite{he2024mallm} & 38.7 & 3.1 & 15.0 & 10.1 & 4.8 & 2.2 & 1.1 \\
    $\text{M}^3$~\cite{cai2024m3} & 47.8 & 3.0 & 16.4 & 16.7 & 6.9 & 2.8 & 1.2 \\
    \midrule
    \bf \method~(Ours) & \bf 53.7 & \bf 11.1 & \bf 19.8 &
    \bf 26.7 & \bf 12.3 & \bf 5.7 & \bf 2.7 \\
    \bottomrule
\end{tabular}
}
\label{tbl:temporalbench}
\vspace{-2mm}
\end{table*}

\subsection{Main Results}
\label{subsec:main_results}
Despite never seeing any manual labels, \method~demonstrates strong performance across all benchmarks. In Tables~\ref{tbl:charades&activity} and~\ref{tbl:tvgbench&rextime&etbench}, \method~surpasses all SFT-based models and achieves first- or second-best performance among RL-based models, which, however, rely on manually labeled data. In Table~\ref{tbl:temporalbench}, we evaluate the proposer on TemporalBench, which reflects the model’s fine-grained captioning capability to describe a sequence of events throughout the video. All models here use the single-pass protocol of the original benchmark. \method~outperforms all prior models across every metric, with particularly large margins on CIDEr and BLEU, despite never being explicitly trained for captioning. Beyond performance, \method~is significantly more data-efficient than existing models, which typically involve data generation and curation pipelines. For instance, VideoChat-R1.5~\cite{yan2025videochatr_1_5} constructs VTTS-80K from 11 public datasets, and ED-VTG~\cite{pramanick2025enrich} collects 136K samples from 8 public datasets, whereas we use only 2.5K raw videos. We also compare our training budget against Time-R1~\cite{wang2025timer1} and ED-VTG~\cite{pramanick2025enrich} in Appendix Table~\ref{appendix:tbl:efficiency}. Overall, these results demonstrate that \method~effectively learns temporal information without manual labels.
\section{Analysis}
\label{sec:analysis}
In this section, we provide in-depth analyses of \method.
\begin{figure}[t]
        \centering
        \includegraphics[width=0.99\linewidth]{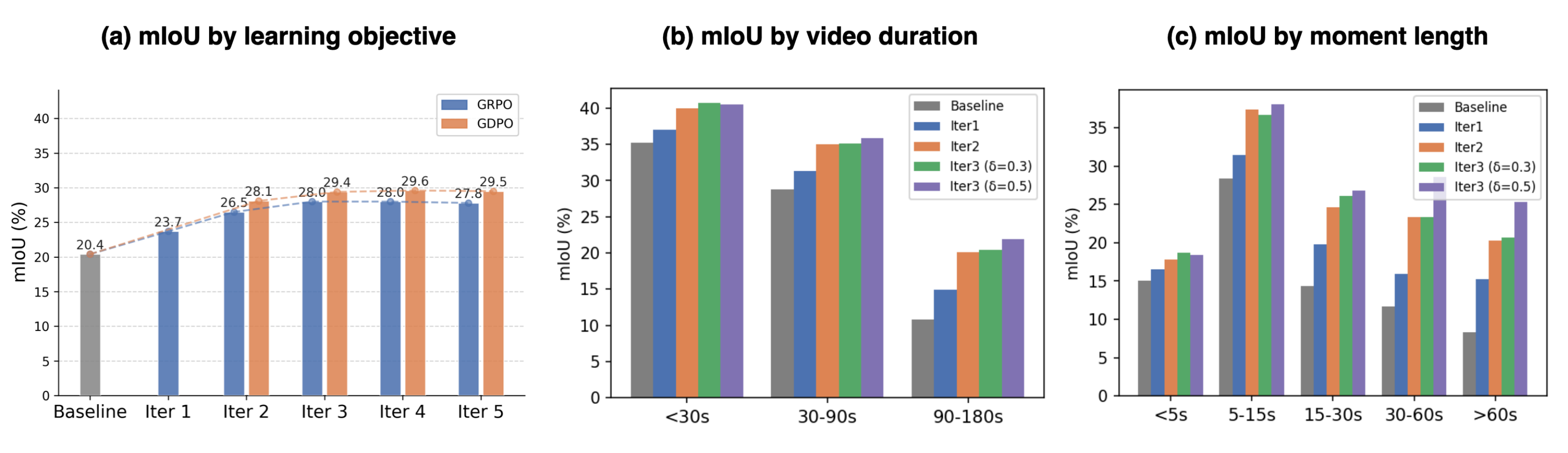}
    \caption{
    \textbf{Improvements across iterations on TVGBench.} (a) shows performance using different learning objectives. (b) and (c) show performance across different video and moment lengths.
    }
    \label{fig:iterations}
\end{figure}
\begin{table*}[t]
\centering
\begin{minipage}{0.5\textwidth}
    \centering
    \caption{
    \textbf{Ablation on data scaling.} FT indicates fine-tuning on Charades-STA training data. In contrast, EvoGround (7.5K) additionally uses 5K raw videos from Charades-STA without manual labels. 
    }
    \label{tbl:scaling}
    \resizebox{\linewidth}{!}{
    \begin{tabular}{lc cccc}
    \toprule
    \multirow{2}{*}{\bf Method} & \multirow{2}{*}{\bf FT} & \multicolumn{4}{c}{\bf Charades-STA} \\
    \cmidrule(lr){3-6} 
     & & R1@0.3 & R1@0.5  & R1@0.7 & mIoU \\
    \midrule  

    Hawkeye~\cite{wang2024hawkeye} & \cmark 
    & 72.5 & 58.3 & 28.8 & 49.3 \\
    
    TimeChat~\cite{timechat} & \cmark
    & - & 46.7 & 23.7 & - \\

    TimeChat-VT~\cite{jung2025consistency} & \cmark
    & - & 58.4 & 34.7 & - \\
    
    ED-VTG~\cite{pramanick2025enrich} & \cmark
    & 78.2 & 62.1 & 35.0 & 52.6 \\
    
    VideoChat-R1~\cite{li2025videochatr1} & \cmark
    & - & \underline{71.7} & \bf 50.2 & \underline{60.8} \\
    
    VideoChat-R1.5~\cite{yan2025videochatr_1_5} & \cmark
    & \bf 82.8 & 71.6 & 48.3 & 60.6 \\

    Time-R1~\cite{wang2025timer1} & \cmark
    & \bf 82.8 & \bf 72.2 & \underline{50.1} & \bf 60.9 \\
    
    \midrule
    \textbf{EvoGround} (\small2.5K) & \xmark 
    & 77.2 & 60.5 & 35.5 & 53.1 \\
    
    \textbf{EvoGround} (\small7.5K) 
    & \xmark 
    & \underline{81.7} & 67.7 & 45.6 & 60.3 \\
    \bottomrule
    \end{tabular}
}
\end{minipage}
\hfill
\begin{minipage}{0.48\textwidth}
    \centering
    \caption{\textbf{Ablation on rewards.} Each reward contributes incrementally, with all rewards combined achieving the best performance.  
    }
    \label{tbl:ablation_on_rewards}
    \resizebox{\linewidth}{!}{
    \resizebox{\linewidth}{!}{
    \begin{tabular}{cccc ccc}
    \toprule
    \multirow{2}{*}{\bf Iter} 
    & \multirow{2}{*}{\bf $R_{\text{fmt}}$} 
    & \multirow{2}{*}{\bf $R_{\text{con}}$} 
    & \multirow{2}{*}{\bf $R_{\text{feed}}$} 
    & \multicolumn{3}{c}{\bf TVGBench} \\
    \cmidrule(lr){5-7}
    & & & & R1@0.3 & R1@0.5 & R1@0.7 \\
    \midrule
    1 & \checkmark & & & 34.0 & 21.2 & 12.0 \\
    \cmidrule(lr){1-7}
    2 & \checkmark & & & 37.0 & 23.8 & 12.5 \\
    2 & \checkmark & \checkmark & & \underline{38.7} & \underline{26.2} & \underline{14.6} \\
    2 & \checkmark & & \checkmark & 38.1 & 26.0 & 14.4 \\
    2 & \checkmark & \checkmark & \checkmark & \bf 40.1 & \bf 27.4 & \bf 14.9 \\
    \cmidrule(lr){1-7}
    3 & \checkmark & & & 37.2 & 23.6 & 13.5 \\
    3 & \checkmark & \checkmark & & 41.7 & 27.3 & \underline{15.0} \\
    3 & \checkmark & & \checkmark & \underline{41.8} & \underline{27.4} & 14.8 \\
    3 & \checkmark & \checkmark & \checkmark & \bf 42.1 & \bf 28.5 & \bf 15.3 \\
    \bottomrule
    \end{tabular}}
    }
\end{minipage}
\end{table*}

\begin{table*}[t]
    \centering
    \caption{\textbf{Co-evolution vs.\ static pseudo-labeling on TVGBench} (R1@0.3 / R1@0.5 / R1@0.7). All rows use the same 2.5K raw videos, backbone and RL objective, matched in total solver update steps. For the static controls (B0--B2), each ``iteration'' is an additional pass over the same fixed data.}
    \label{tbl:coevolution}
    \resizebox{0.95\linewidth}{!}{
    \begin{tabular}{l ccc}
    \toprule
    \textbf{Method} & \textbf{Iter 1} & \textbf{Iter 2} & \textbf{Iter 3} \\
    \midrule
    B0 (random moment + captioning)        & 30.2 / 19.8 / 10.8 & 32.4 / 20.2 / 10.9 & 33.8 / 21.6 / 11.3 \\
    B1 (one-shot pseudo-label)             & 31.3 / 20.2 / 10.6 & 32.0 / 20.3 / 11.0 & 32.3 / 20.5 / 11.1 \\
    B2 (frozen iter-1 bootstrap proposer)  & 34.0 / 21.2 / 12.0 & 35.2 / 24.8 / 12.5 & 37.3 / 25.5 / 13.3 \\
    \midrule
    \bf \method~(full co-evolution)        & 34.0 / 21.2 / 12.0 & \bf 40.1 / 27.4 / 14.9 & \bf 42.1 / 28.5 / 15.3 \\
    \bottomrule
    \end{tabular}
    }
\end{table*}

\textbf{GDPO outperforms GRPO.}
Figure~\ref{fig:iterations} (a) reports the performance using different RL learning objectives across iterations. GDPO consistently outperforms GRPO across all iterations and already surpasses GRPO's best performance at the second iteration, achieving faster convergence. These results support our claim that GDPO is well-suited for our multi-reward settings.

\textbf{Increasing $\delta$ shifts the model’s focus toward more accurate grounding.} 
We compare the cases where $\delta$ is increased to 0.5 versus kept at 0.3. In Figure~\ref{fig:iterations} (b) and (c), $\delta = 0.5$ shows a more balanced improvement than $\delta=0.3$ across different video durations and moment lengths. Specifically, $\delta=0.3$ yields only marginal improvement over the previous iteration, and its effectiveness diminishes particularly for longer videos and moments. This suggests that increasing $\delta$ encourages the proposer to foster a more robust solver. This does not bias toward trivially easy samples, as the improvement is most pronounced for longer videos and moments where grounding is inherently harder.

\textbf{Co-evolution, not repeated self-training, drives the gains.}
Because the solver trains on proposer outputs, \method~might reduce to one round of RL on filtered self-generated data. We test this with three controls on the same videos, backbone and objective, matched in solver update steps: random moments captioned by the frozen base model (B0), one-shot pseudo-labels from the frozen base model (B1), and a proposer frozen after one bootstrap iteration (B2), where each later iteration is another pass over the same fixed data. In Table~\ref{tbl:coevolution} the static pipelines plateau ($+3.3$ R1@0.3 for B2) while \method~gains $+8.1$, so the benefit comes from updating the generator, not from further solver training.

\textbf{Adding more videos restarts the loop.} Figure~\ref{fig:iterations} (a) shows that gains plateau by iteration 3. With only 2.5K raw videos, the proposer and solver have already learned what they can from this small pool, and later iterations mostly recycle similar pairs. In Table~\ref{tbl:scaling}, training~\method~on 5K additional raw videos from Charades-STA (7.5K total, no overlap) for one extra iteration yields further gains, remaining competitive with models fine-tuned on human annotations. Appendix Table~\ref{appendix:tbl:upscale} shows smaller but positive transfer to other benchmarks.

\begin{figure}[t]
        \centering
        \includegraphics[width=0.99\linewidth]{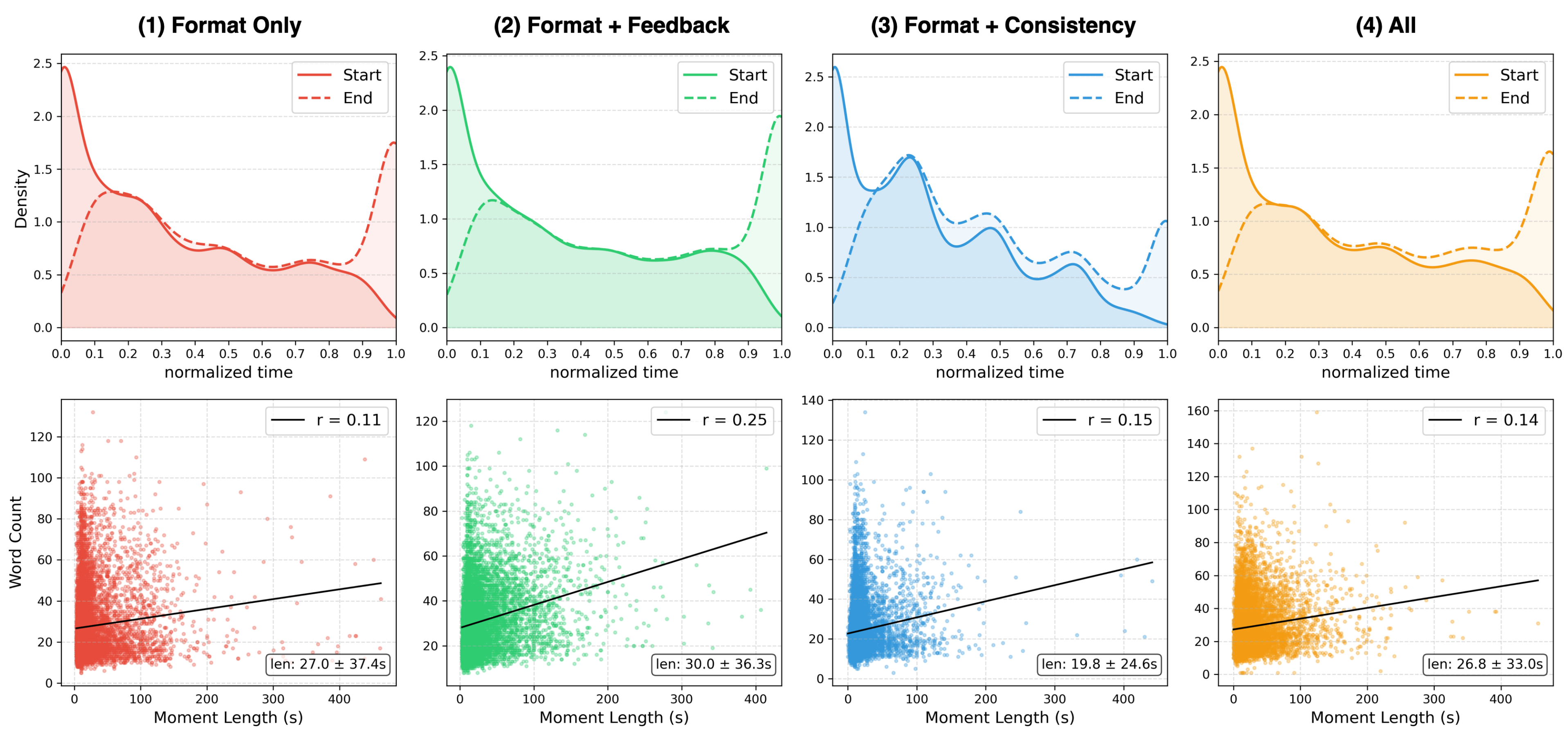}
    \caption{
    \textbf{Generated data distributions across different reward configurations.} 
    \textit{Top:} Kernel density estimates of normalized start and end times of the moment, shown as solid and dashed lines, respectively. All times are normalized by video duration. \textit{Bottom:} We report the correlation ($r$) between query and moment length, along with the mean $\pm$ standard deviation of moment lengths. 
    }
    \label{fig:histogram}
\end{figure}

\textbf{Each reward plays a distinct role.} 
We optimize our model under different reward configurations: (1) format only, (2) format + feedback, (3) format + consistency, and (4) all rewards, and visualize their moment distributions and query--moment length correlation ($r$) in Figure~\ref{fig:histogram}, with performance in Table~\ref{tbl:ablation_on_rewards}. Each reward addresses a specific failure mode of the proposer, jointly preventing collapse to trivial outputs. The format reward enforces structural validity, but formatting alone yields the worst performance, suggesting the proposer collapses into biased, repetitive data without quality constraints. The feedback reward enriches query descriptiveness ($r$: 0.11~$\rightarrow$~0.25), though $r$ is a diagnostic signal rather than a quality metric: a longer query does not necessarily describe the moment better. The consistency reward sharpens the moment distribution, producing tighter temporal moments (27.0~$\pm$~37.4~$\rightarrow$~19.8~$\pm$~24.6) and reducing extension toward the video end. The tri-modal start time distribution reflects the consistency reward's tendency to anchor moments at discriminative positions, avoiding overlap with other candidates. Combining all rewards moderates $r$ to 0.14 and yields the balanced distribution and best performance overall. 

\textbf{Gains transfer across backbone scales and families.} Across four backbones spanning 3B--8B parameters and two families (\ie, QwenVL and InternVL), \method~delivers nearly identical gains and benefits from larger capacity, with Qwen3-VL-8B~\cite{bai2025qwen3} reaching the highest absolute scores. We will explore more aggressive scaling in future work.
\begin{table*}[t]
\centering
\begin{minipage}{0.56\textwidth}
    \centering
    \caption{
    \textbf{Ablation on backbones.} \textcolor{gray}{Gray} columns show initial performance, and \textcolor{blue}{blue} indicates improvements from \method. 
    Gains are stable across backbone size and family, suggesting the framework rather than backbone-specific behavior drives improvement.
    }
    \label{tbl:backbone_ablation}
    \resizebox{\linewidth}{!}{
    \begin{tabular}{
    l 
    >{\columncolor{gray!15}}c c 
    >{\columncolor{gray!15}}c c 
    >{\columncolor{gray!15}}c c}
    \toprule
    \multirow{2}{*}{\bf Backbone} 
    & \multicolumn{6}{c}{\bf TVGBench} \\
    \cmidrule(lr){2-7}
    & \multicolumn{2}{c}{R1@0.3} 
    & \multicolumn{2}{c}{R1@0.5} 
    & \multicolumn{2}{c}{R1@0.7} \\
    \midrule
    
    Qwen2.5-VL-3B~\cite{bai2025qwen2} 
    & 17.5 & \bf 31.3 {\footnotesize\textcolor{blue}{(+13.8$\uparrow$)}} 
    & 10.0 & \bf 19.4 {\footnotesize\textcolor{blue}{(+9.4$\uparrow$)}} 
    & 5.2 & \bf 9.9 {\footnotesize\textcolor{blue}{(+4.7$\uparrow$)}} \\
    Intern3.5-VL-4B~\cite{wang2025internvl3} 
    & 14.2 & \bf 30.6 {\footnotesize\textcolor{blue}{(+16.4$\uparrow$)}} 
    & 8.5 & \bf 18.1 {\footnotesize\textcolor{blue}{(+9.6$\uparrow$)}}
    & 4.0 & \bf 9.6 {\footnotesize\textcolor{blue}{(+5.6$\uparrow$)}} \\
    Qwen2.5-VL-7B~\cite{bai2025qwen2} 
    & 28.1 & \bf 42.1 {\footnotesize\textcolor{blue}{(+14.0$\uparrow$)}} 
    & 19.5 & \bf 28.5 {\footnotesize\textcolor{blue}{(+9.0$\uparrow$)}} 
    & 10.5 & \bf 15.3 {\footnotesize\textcolor{blue}{(+4.8$\uparrow$)}} \\
    Qwen3-VL-8B~\cite{bai2025qwen3} 
    & 30.2 & \bf 44.6 {\footnotesize\textcolor{blue}{(+14.4$\uparrow$)}} 
    & 20.8 & \bf 30.1 {\footnotesize\textcolor{blue}{(+9.3$\uparrow$)}} 
    & 10.5 & \bf 17.2 {\footnotesize\textcolor{blue}{(+6.7$\uparrow$)}} \\
    
    \bottomrule
    \end{tabular}
    }
\end{minipage}
\hfill
\begin{minipage}{0.41\textwidth}
    \centering
    \caption{
    \textbf{Dense video captioning results.} \method~remains competitive with existing models. FT denotes training on dense captions in ActivityNet-Captions.
    }
    \label{tbl:dvc}
    \resizebox{\linewidth}{!}{
    \begin{tabular}{lc cc}
    \toprule
    \multirow{2}{*}{\textbf{Method}} & \multirow{2}{*}{\textbf{FT}} & \multicolumn{2}{c}{\textbf{ActivityNet-Captions}} \\
    \cmidrule(lr){3-4} & & SODA\_c & METEOR \\
    \midrule
    Momentor~\cite{qian2024momentor} & \xmark & 2.3 & 4.7 \\
    VTimeLLM~\cite{vtimellm} & \cmark & 5.8 & \underline{6.8} \\
    TRACE~\cite{guo2024trace} & \cmark & \bf 6.4 & 6.0 \\
    Grounded-VideoLLM~\cite{wang2024groundedvideollm} & \cmark & \underline{6.2} & 6.4 \\ \midrule
    \bf \method~(Ours) & \xmark & 4.1 & \bf 7.0 \\
    \bottomrule
    \end{tabular}
    }
\end{minipage}
\end{table*}

\textbf{\method~generalizes beyond grounding.} We evaluate the proposer on dense video captioning~\cite{krishna2017dense}. Table~\ref{tbl:dvc} reports results on ActivityNet-Captions, showing that \method~achieves the highest METEOR score among all models. Despite a lower SODA\_c score, it remains competitive without involving any caption data, unlike prior methods fine-tuned on human caption supervision. We further evaluate the solver on representative VideoQA benchmarks, Video-MME~\cite{fu2024videomme} and TempCompass~\cite{liu2024tempcompass}, in Appendix Table~\ref{appendix:tbl:videoqa}, where \method~achieves slight improvements over the backbone without disrupting its original QA capabilities.

\definecolor{skyblue}{rgb}{0.53, 0.81, 0.92} 

\begin{table*}[t]
\centering
\caption{\textbf{Sensitivity of the consistency reward to encoder choice.} We report reward statistics (mean $\pm$ standard deviation) alongside
TVGBench performance using SigLIP-2 and CLIP ViT-B/32. The last row reports cross-encoder Spearman correlations ($\rho$) for the reward components and final reward.}
\label{tab:encoder-consistency}
\resizebox{\linewidth}{!}{%
\begin{tabular}{lccccc}
\toprule
\textbf{Encoder}
& \textbf{Similarity ($\mu$)}
& \textbf{Inter-consistency}
& \textbf{Intra-consistency}
& \textbf{Final reward} 
& \bf mIoU \\
\midrule
SigLIP-2~\cite{tschannen2025siglip}
& $0.015 \pm 0.023$
& $0.206 \pm 0.027$
& $0.732 \pm 0.117$
& $0.150 \pm 0.028$ 
& 29.1 \\
CLIP ViT-B/32~\cite{radford2021learning}
& $0.270 \pm 0.042$
& $0.233 \pm 0.080$
& $0.556 \pm 0.198$
& $0.127 \pm 0.060$ 
& 29.4 \\
\midrule
\rowcolor{skyblue!30}
\textcolor{blue}{Spearman correlations ($\rho$)}
& \textcolor{blue}{0.09}
& \textcolor{blue}{0.12}
& \textcolor{blue}{0.50}
& \textcolor{blue}{0.36}
& - \\
\bottomrule
\end{tabular}%
}
\end{table*}
\textbf{The consistency reward remains comparable in scale across encoders.}
We investigate the effect of different visual encoders on consistency reward computation. In Table~\ref{tab:encoder-consistency}, we sample 500 query--moment pairs from TimeRFT and evaluate the components of the consistency reward using SigLIP-2 and CLIP ViT-B/32. Although the mean similarities differ by more than an order of magnitude (0.015 vs.\ 0.270), both the mean consistency rewards (0.150 vs.\ 0.127) and performance (29.1 vs.\ 29.4) remain comparable. This suggests that the softmax-based inter-term and bounded intra-consistency help mitigate differences in similarity scales across visual encoders. Together with their comparable downstream performance, this suggests the consistency reward can be computed stably with different visual encoders, although the softmax temperature $\tau$ may require recalibration per encoder.
\section{Conclusion}
\label{sec:conclusion}
We presented \method, a framework of coupled self-evolving agents that learns temporal grounding from raw videos without any task-specific grounding labels. Through a proposer-solver loop driven entirely by reinforcement learning, the two agents iteratively challenge and improve each other, yielding a system that matches or outperforms most fully supervised models across multiple benchmarks for both grounding and captioning. We hope this work inspires a broader shift toward annotation-free video understanding systems that learn directly from the raw visual world.

\noindent \textbf{Limitations.} 
Despite the effectiveness of \method, several challenges remain. First, it becomes costly when handling ultra-long videos, as it generates a large number of query--moment pairs and needs the solver's feedback for each pair. Second, gains are front-loaded and saturate by iteration 3 on a fixed corpus. Scaling the corpus restarts the loop, but we have not verified beyond 7.5K videos that the ceiling is imposed by the corpus rather than the mechanism (see Appendix~\ref{appendix:subsec:discussion on training data}). Third, the loop is annotation-free but not prior-free: the consistency reward uses a general-purpose image--text encoder, and a non-web-aligned encoder would be the cleanest test of where the signal originates. Fourth, the solver's reward treats the proposer's moment as ground truth, so the objective is self-referential, mitigated but not replaced by our external-judge audit.

\newpage
\section*{Acknowledgements and Disclosure of Funding}
This work was partly supported by the IITP (RS-2021-II212068-AIHub/10\%, RS-2021-II211343-GSAI/15\%, RS-2022-II220951-LBA/15\%, RS-2022-II220953-PICA/20\%), NRF (RS-2024-00353991-SPARC/20\%, RS-2023-00274280-HEI/10\%), and KEIT (RS-2024-00423940/10\%) grant funded by the Korean government

\noindent \textbf{Competing interests.} The authors declare no competing interests. No third-party funding beyond the grants listed above supported this work during the preceding 36 months, and the authors have no financial relationships outside the submitted work that could be perceived as influencing it.
\bibliography{main}
\bibliographystyle{ieeetr}

\newpage
\appendix
\section*{Appendix}
In this section, we provide details that are not included in the main manuscript due to the page limit. 

\section{Details on~\method}
\label{appendix:sec:model details}
\subsection{Group reward-Decoupled Normalization Policy Optimization (GDPO)}
\label{appendix:subsec:gdpo} 
We describe GDPO~\cite{liu2026gdpo}, an RL optimization algorithm derived from GRPO~\cite{guo2025deepseek}. We first briefly review GRPO. Given a prompt $p_i$, the model samples $G$ candidate responses $\{o_1, \ldots, o_G\}$ from the behavior policy $\pi_{\theta_{\text{old}}}$. GRPO computes a group-normalized advantage $A_{\text{sum}}^{(i,j)}$ by first summing reward components into $R_{\text{sum}}^{(i,j)}$ and then normalizing:
\begin{equation}
    A_{\text{sum}}^{(i,j)} = \frac{R_{\text{sum}}^{(i,j)} - 
    \operatorname{mean}\{R_{\text{sum}}^{(i,1)},\ldots,R_{\text{sum}}^{(i,G)}\}}
    {\operatorname{std}\{R_{\text{sum}}^{(i,1)},\ldots,R_{\text{sum}}^{(i,G)}\}}, \quad
    R_{\text{sum}}^{(i,j)} = R_1^{(i,j)} + \cdots + R_k^{(i,j)}.
    \label{eq:GRPO advantage}
\end{equation}
GRPO then optimizes the policy to favor responses with higher relative advantage. However, such naively summing rewards overlooks differences in their difficulty and scale, often biasing optimization toward easier objectives while neglecting more challenging ones~\cite{skalse2022defining}. GDPO addresses this by reversing the order: it normalizes each reward component \emph{individually before} aggregation, rather than after. Specifically, for each reward $R_k$, GDPO computes a per-reward normalized advantage $A_k^{(i,j)}$:
\begin{equation}
    A_k^{(i,j)} = \frac{R_k^{(i,j)} - \operatorname{mean}\{R_k^{(i,1)},
    \ldots,R_k^{(i,G)}\}}{\operatorname{std}\{R_k^{(i,1)},\ldots,
    R_k^{(i,G)}\}},
\end{equation}
and then combines them with defined weights before a final normalization:
\begin{gather}
    A_{\text{sum}}^{(i,j)} = w_1 A_{1}^{(i,j)} + w_2 A_{2}^{(i,j)} 
    + \cdots + w_k A_{k}^{(i,j)}, \\
    \hat{A}_{\text{sum}}^{(i,j)} = \frac{A_{\text{sum}}^{(i,j)} - 
    \operatorname{mean}\{A_{\text{sum}}^{(i',j')}|i' \in \mathcal{B}, j'=1,...,G \}}
    {\operatorname{std}\{A_{\text{sum}}^{(i',j')}|i' \in \mathcal{B}, j'=1,...,G\}},
\end{gather}
which applies batch-wise ($\mathcal{B}$) advantage normalization. This design enables prioritization of more critical objectives and is particularly suitable for multi-reward settings. We refer the reader to~\cite{guo2025deepseek} and~\cite{liu2026gdpo} for the full derivations of GRPO and GDPO, respectively.

\subsection{Additional Implementation Details.} 
\label{appendix:subsec:additional implementation details}
\noindent \textbf{Hyperparameters.} For both agents, we use the AdamW optimizer with a learning rate of 1e-6. We set the number of rollouts (G) to 2. We use a batch size of 1 with gradient accumulation of 4. We set the max token lengths to 512 for the proposer and to 200 for the solver. This considers their different role, as the proposer likely outputs longer answers than the solver since we require both natural language queries and corresponding timestamps. For the proposer, we set $\tau_{\text{length}}=3$ and $\gamma=30$.  No labeled validation set is used at any stage, since the setting assumes no annotations, and no benchmark score was consulted for any hyperparameter decision. The $\delta$ schedule ($0 \rightarrow 0.3 \rightarrow 0.5$) was chosen from training-internal signals alone: with $\delta$ fixed at 0.3 the accuracy and feedback reward curves stagnated, and raising $\delta$ across iterations restored a healthy progression (cf.\ Figure~\ref{fig:rewards}). The reward weights and $\tau_{\text{length}}$ were set to conventional values without tuning, and identical hyperparameters are used across all five benchmarks.

\begin{figure}[t]
        \centering
        \includegraphics[width=0.99\linewidth]{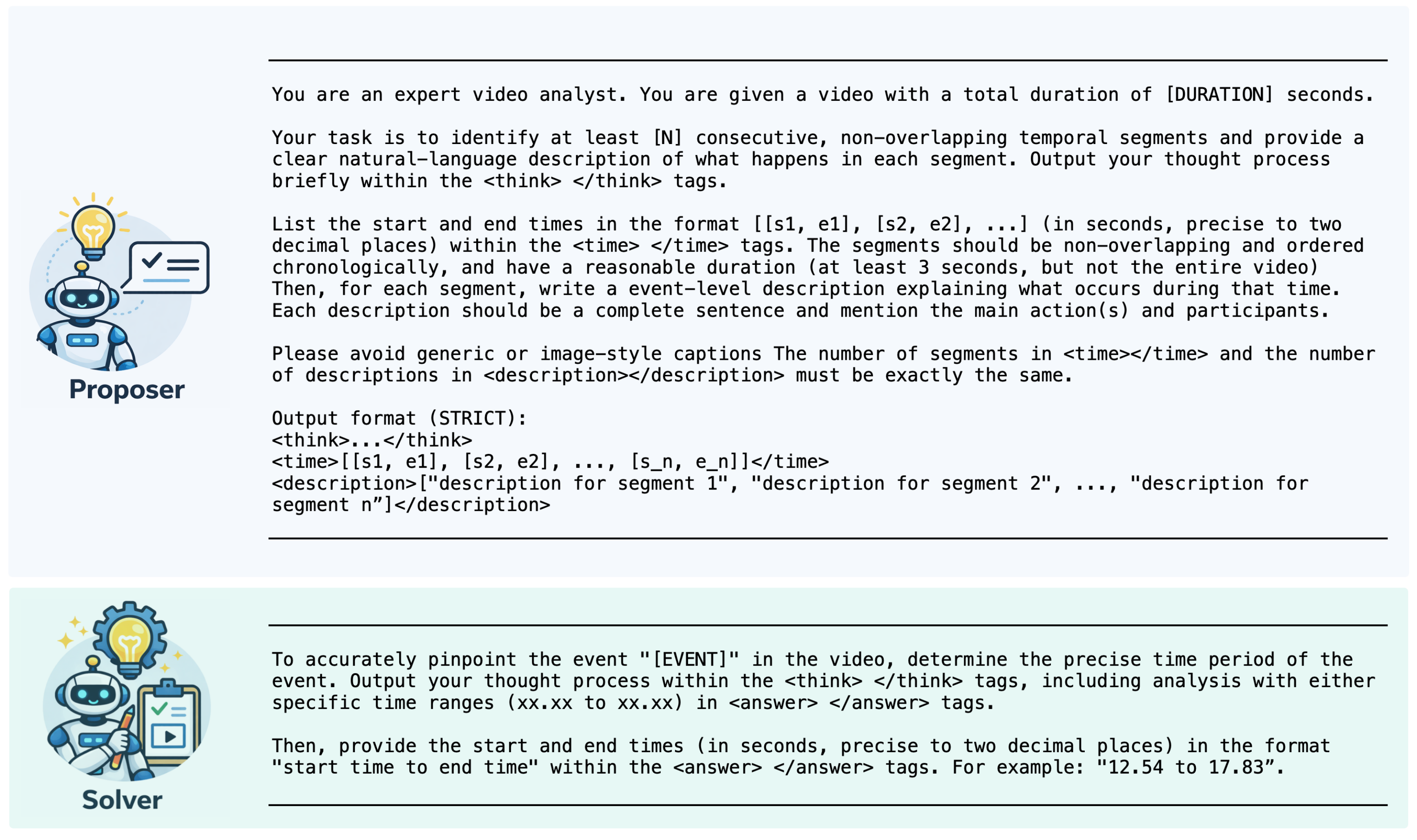}
    \caption{
    \textbf{Prompt designs of \method.} We show the prompts used for the proposer (top) and solver (bottom). The proposer is instructed to generate consecutive, non-overlapping query--moment pairs from a raw video, while the solver is instructed to localize a given query within the video.}
    \label{appendix:fig:prompt}
\end{figure}

\noindent \textbf{Full prompts.} In Figure~\ref{appendix:fig:prompt}, we provide a designed prompt for the proposer and solver, respectively. The proposer is encouraged to generate query--moment samples under valid conditions, and the solver aims to temporally ground the video from the given query sentence. Here, we set $N$ to 4 for the proposer. If the proposer generates valid query--moment pairs less than $N$, the remaining entries are padded with zero moments to penalize insufficient sampling and encourage the generation of multiple pairs. For instance, if the proposer generates only one valid pair (\eg, covering the entire video), the format reward is reduced to 0.25 since the remaining three pairs are assigned 0.

\noindent \textbf{Memory peak.} During trainnig, the peak memory is $\sim$60\,GB/GPU (proposer) and $\sim$30\,GB/GPU (solver).

\subsection{Details on the Reward Designs}
\label{appendix:subsec:reward designs}
\noindent \textbf{SigLIP-2.} SigLIP-2~\cite{tschannen2025siglip} is a widely used vision-language encoder that extends the success of the original SigLIP with enhanced pre-training objectives. We use the 0.4B-scale model, which is lightweight and incurs negligible computations, for the consistency reward during training.

\noindent \textbf{Intra-Consistency.}
In Equation~\ref{eq:intra-consistency}, the exponential formulation is designed to map the variance of frame-level similarities to a sharply discriminative score. Since similarity values lie in $[-1, 1]$, the resulting variance $\sigma_n$ is bounded within a limited range. In the ideal case, where all frames within a moment are perfectly aligned with the query, the variance becomes zero, yielding $\exp(0)=1$. In contrast, highly inconsistent moments exhibit large fluctuations in similarity (\ie, alternating between 1 and $-1$), resulting in $\sigma_n \approx 1$. Such cases are mapped to a near-zero value via the exponential function, $\exp(-1 \times 30) \approx 9.3 \times 10^{-14}$. This design ensures that well-aligned moments are preserved with high scores, while even moderate inconsistencies are strongly penalized, providing a clear separation between coherent and incoherent moment predictions. Additionally, we determine $|m_n|$ in Equation~\ref{eq:intra-consistency} as:
\begin{equation}
    |m_n| = \operatorname{clip}\big( (e_n - s_n)\cdot \rho, T_{\min}, T_{\max}),
\end{equation}
where $\rho = 1$, $T_{\min} = 8$, and $T_{\max} = 32$. This ensures that longer moments are represented with more sampled frames.

\noindent \textbf{Variants of Inter-Consistency.} We consider two variants of inter-consistency: a hard ranking-based formulation and an exponential formulation:
\begin{equation}
\text{Inter}_n^{\text{hard}} = 
\frac{1}{N} \sum_{j=1}^{N} 
\mathbb{I}\big( \mu_n \geq \mu_{n,j} \big),
\end{equation}
which measures the fraction of candidates that the $n$-th sample outperforms, capturing its relative ranking within the group. We also design the function with an exponential formulation:
\begin{equation}
    \text{Inter}_n^{\text{exp}} =
    \begin{cases}
    1 - \exp\left(
    -\gamma_{\text{inter}} \cdot
    \frac{1}{N-1} \sum_{j \neq n} \max(0,\, \mu_n - \mu_{n,j})
    \right), & \text{if } N > 1, \\
    0, & \text{otherwise}.
    \end{cases}
\end{equation}
which provides a smooth and more sensitive measure of relative superiority, where larger margins between $\mu_n$ and other candidates lead to exponentially higher scores. We set $\gamma_{\text{inter}}$ to 30. In Table~\ref{appendix:tbl:inter-consistency}, we compare the performance of applying these inter-consistency variants and confirm that computing the inter-consistency using a softmax demonstrates the best performance. This is likely due to its smooth and more discriminative treatment of relative differences among candidates compared to other variants.

\begin{table*}[t]
\centering
\begin{minipage}{0.42\textwidth}
    \centering
    \caption{\textbf{Ablation on the variants of inter-consistency.} 
    }
    \resizebox{\linewidth}{!}{
    \begin{tabular}{l ccc}
    \toprule
    \multirow{2}{*}{\bf Intra} & \multicolumn{3}{c}{\bf TVGBench} \\
    \cmidrule{2-4} & R1@0.3 & R1@0.5  & R1@0.7 \\
    \midrule
    hard & 41.7 & 27.3 & 15.0 \\
    exp &  40.8 & 26.9 & 14.3 \\
    softmax & \bf 42.1 & \bf 28.5 & \bf 15.3 \\
    \bottomrule
    \end{tabular}
    }
    \label{appendix:tbl:inter-consistency}
\end{minipage}
\hfill
    \begin{minipage}{0.54\textwidth}
    \centering
    \caption{\textbf{Results on Video-MME and TempCompass.} * denotes our reproduced results.
    }
    \label{appendix:tbl:videoqa}
    \resizebox{\linewidth}{!}{
    \begin{tabular}{l cccc}
    \toprule
    \multirow{2}{*}{\bf Method} & \bf Video-MME & \bf TempCompass \\
    \cmidrule{2-2} \cmidrule{3-3} & w/o subs & multi-choice QA \\
    \midrule
    Qwen2.5-VL-7B*~\cite{bai2025qwen2} & \underline{61.4} & \underline{71.3} \\
    Time-R1*~\cite{wang2025timer1} & 61.2 & 71.2 \\ \midrule
    \bf \method~(Ours) & \bf 62.3 & \bf 71.5 \\
    \bottomrule
    \end{tabular}
    }
\end{minipage}
\end{table*}
\begin{figure}[t]
        \centering
        \includegraphics[width=0.99\linewidth]{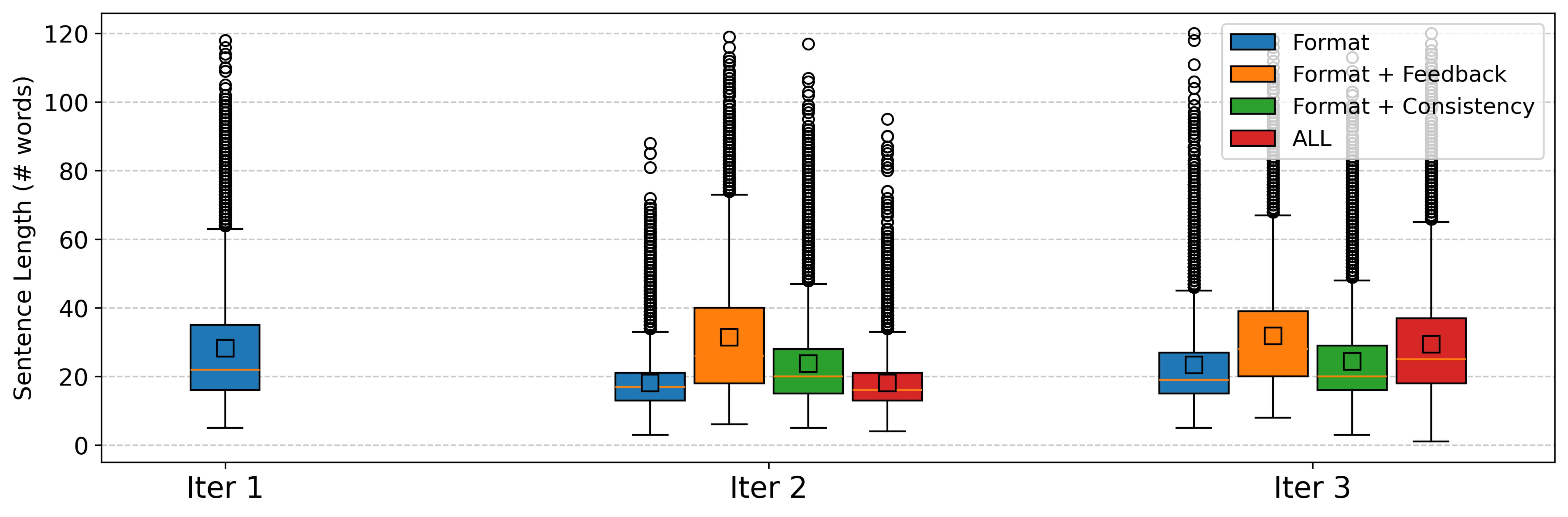}
    \caption{
    \textbf{Query length distribution across iterations.} We visualize the query length distributions across different reward configurations and iterations. As previously discussed in Section~\ref{sec:analysis}, the feedback reward increases the length of queries compared to others.
    }
    \label{appendix:fig:query_length_distribution}
\end{figure}
\begin{figure}[t]
        \centering
        \includegraphics[width=0.99\linewidth]{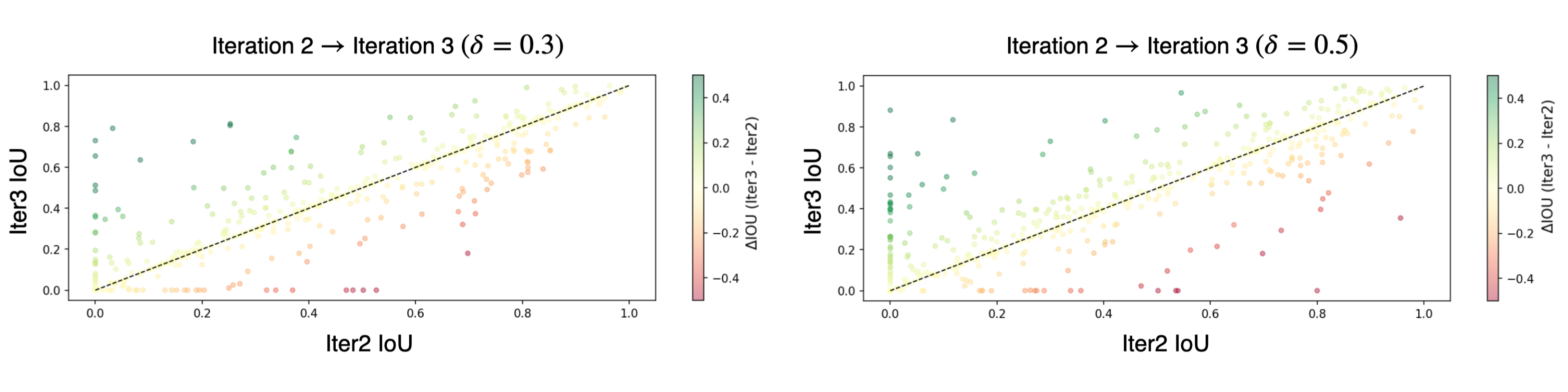}
    \caption{
    \textbf{Per-sample IoU improvements under different thresholds $\delta$.} The dashed line represents no change, and the upper-left regions represent improved cases.
    }
    \label{appendix:fig:per_sample}
\end{figure}
\begin{table*}[t]
    \centering
    \caption{\textbf{Results on Charades-STA, ActivityNet, and TVGBench.}}
    \label{appendix:tbl:upscale}
    
    \resizebox{\linewidth}{!}{
    \begin{tabular}{l cccc cccc cccc}
    \toprule
    \multirow{2}{*}{\textbf{Method}} & \multicolumn{4}{c}{\textbf{Charades-STA}} & \multicolumn{4}{c}{\textbf{ActivityNet-Captions}} & \multicolumn{4}{c}{\textbf{TVGBench}} \\
    \cmidrule(lr){2-5} \cmidrule(lr){6-9} \cmidrule(lr){10-13}
     & R1@0.3 & R1@0.5 & R1@0.7 & mIoU 
     & R1@0.3 & R1@0.5 & R1@0.7 & mIoU 
     & R1@0.3 & R1@0.5 & R1@0.7 & mIoU \\ 
     \midrule
     \bf \method~(\small2.5K) 
    & 77.2 & 60.5 & 35.5 & 53.1
    & 62.9 & 43.6 & 25.0 & 43.9 
    & 42.1 & 28.5 & 15.3 & 29.4
    \\
     \bf \method~(\small7.5K) & 81.7 & 67.7 & 45.6 & 60.3 & 64.8 & 44.3 & 25.8 & 44.7 & 41.7 & 28.6 & 15.1 & 29.5 \\

    \rowcolor{blue!10} \textcolor{blue}{$\Delta$} 
    & \textcolor{blue}{+4.5} & \textcolor{blue}{+7.2} & \textcolor{blue}{+10.1} & \textcolor{blue}{+7.2} 
    & \textcolor{blue}{+1.9} & \textcolor{blue}{+0.7} & \textcolor{blue}{+0.8} & \textcolor{blue}{+0.8} 
    & \textcolor{red}{-0.4} & \textcolor{blue}{+0.1} & \textcolor{red}{-0.2} & \textcolor{blue}{+0.1} \\

    \bottomrule
    \end{tabular}
    }
\end{table*}

\subsection{Impact of Training Data and Comparison with Time-R1} 
\label{appendix:subsec:discussion on training data}
A potential concern is that the performance of \method~may be influenced by the underlying video corpus used for training. We acknowledge that the choice of data can affect model performance. To control for this factor, we note that both \method~and Time-R1~\cite{wang2025timer1} are trained on the same underlying video corpus (\ie, TimeRFT). Therefore, any performance differences cannot be attributed solely to dataset selection. 

Despite sharing the same video corpus, \method~differs from Time-R1 in several aspects. First, \method~does not rely on any human-annotated data, whereas Time-R1 is trained with labeled supervision. Second, \method~does not require a cold-start fine-tuning stage or the construction of curated chain-of-thought (CoT)~\cite{wei2022chain} examples, both of which are involved in Time-R1 to guide training. Time-R1 further requires dynamic hard sampling during its training that mines hard samples (\ie, low-IoU samples) on a curated dataset for each training epoch. In contrast, \method~learns directly from raw videos through a self-evolving proposer–solver loop driven by RL.

Beyond, \method~is more data-efficient than existing models, achieving strong performance with only a small amount of raw video data. As mentioned in Section~\ref{subsec:main_results}, existing models typically rely on multi-stage training pipelines with curated datasets tailored for each stage. For example, VTimeLLM~\cite{vtimellm} adopts a three-stage framework (feature alignment, boundary perception, and instruction tuning), while Grounded-VideoLLM~\cite{wang2024groundedvideollm} employs a similar multi-stage pipeline that involves large-scale caption supervision (\ie, 1.28M video-caption pairs) in early stages. 

Taken together, these differences suggest that the improvement of \method~stems from the proposed self-evolution framework rather than reliance on curated video quality or dataset-specific advantages. While the underlying corpus still plays a role, our results demonstrate that strong temporal grounding performance can be achieved without annotated data or additional manual curation.

\section{Further Analysis}
\label{appendix:sec:additional_analysis}
\noindent \textbf{Results on VideoQA benchmarks.} We evaluate~\method~on representative video question answering (QA) benchmarks, including Video-MME~\cite{fu2024videomme} and TempCompass~\cite{liu2024tempcompass} in Table~\ref{appendix:tbl:videoqa}. We confirm that \method~does not disrupt the backbone (\ie, Qwen2.5-VL-7B) and demonstrates a slight improvement.

\noindent \textbf{Query length distributions across iterations and different reward configurations.} In Figure~\ref{appendix:fig:query_length_distribution}, we visualize the query length distribution for each reward configuration across iterations. As previously mentioned, the feedback reward enriches query descriptiveness, showing the highest average query length.

\noindent \textbf{Per-sample comparison across different $\delta$.} In Figure~\ref{appendix:fig:per_sample}, we compare per-sample improvements across different $\delta$ on TVGBench. $\delta = 0.5$ results in a denser distribution in the high-IoU region (\ie, upper-left), compared to $\delta = 0.3$, along with higher average improvements (+1.29\% vs. +0.21\%). 

\begin{figure}[t]
        \centering
        \includegraphics[width=0.99\linewidth]{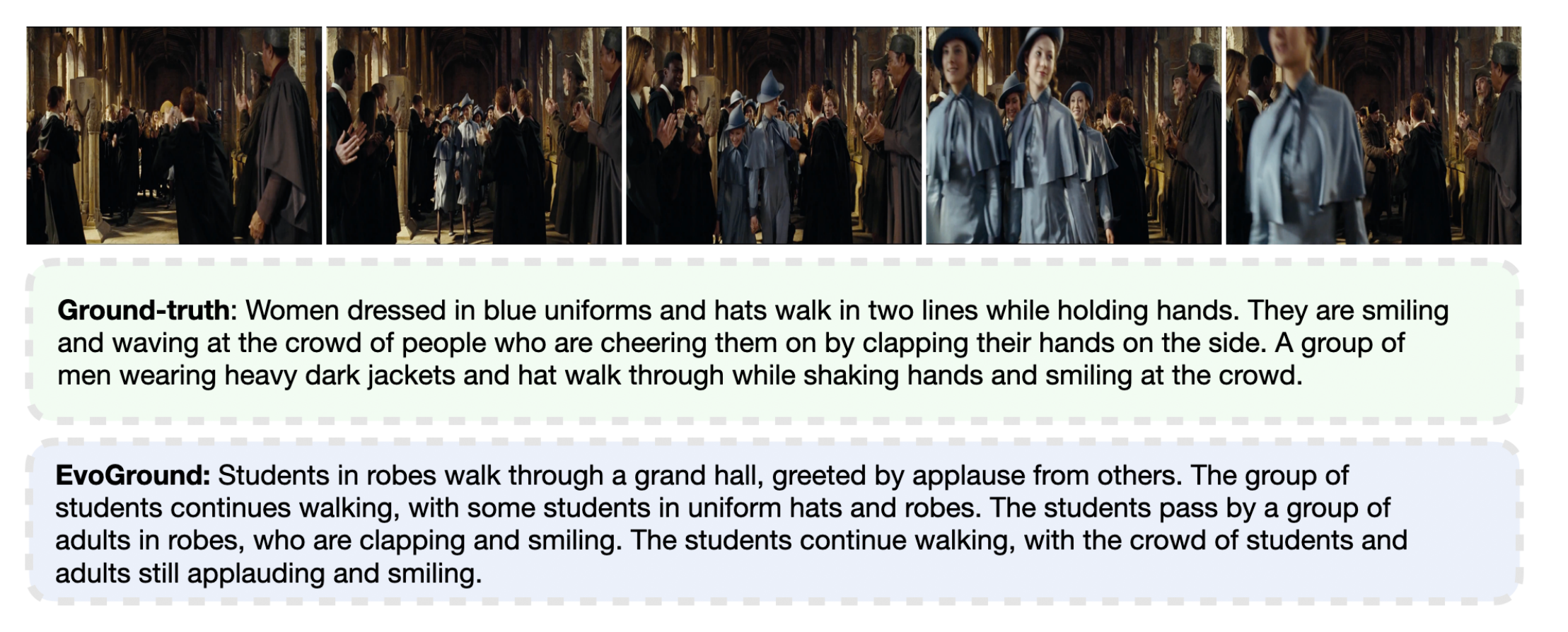}
    \caption{
    \textbf{Captioning results on TemporalBench.} 
    }
    \label{appendix:fig:temporalbench}
\end{figure}

\noindent \textbf{Evaluation on TemporalBench.} As shown in Figure~\ref{appendix:fig:temporalbench}, to evaluate on TemporalBench~\cite{cai2024temporalbench}, we first generate query--moment pairs from given videos and concatenate them into a single paragraph. We then compare it with the ground-truth sentence and confirm that \method~effectively captures consecutive video events.

\begin{figure}[t]
        \centering
        \includegraphics[width=0.99\linewidth]{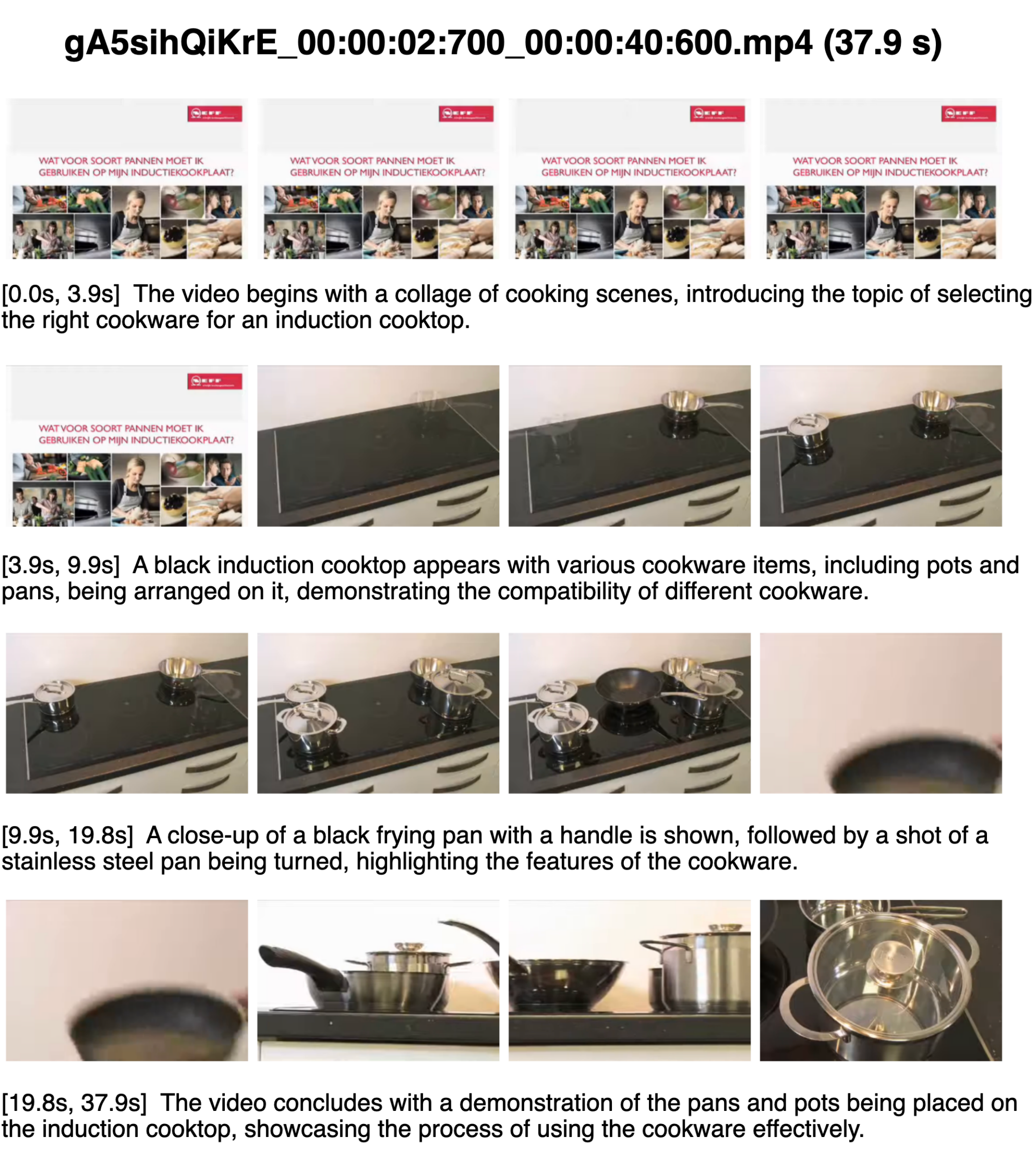}
    \caption{
    \textbf{Generated query--moment pairs from \method.} 
    }
    \label{appendix:fig:samples}
\end{figure}

\noindent \textbf{Visualization of generated data.} We also visualize generated data from the proposer at the third iteration in Figure~\ref{appendix:fig:samples}. Each query is well-aligned with the corresponding frames within the moment and densely covers throughout the video. 

\noindent \textbf{Results of \method~with scaling up video data.} Extending the results in Table~\ref{tbl:scaling}, we report the performance of \method~(7.5K) on additional benchmarks including Charades-STA in Table~\ref{appendix:tbl:upscale}. While we observe some improvements across benchmarks and metrics, the gains are modest, particularly on out-of-distribution datasets such as TVGBench and ActivityNet-Captions. This is expected, as the additional training videos are sourced from Charades-STA, which primarily consists of short videos. Consequently, incorporating such data helps the model better handle similar domains, but is less effective for broader generalization.

\begingroup\color{blue}
\section{Additional Experiments}
\subsection{Generated pairs become more faithful across iterations}
We evaluate both EvoGround-generated query--moment pairs and human-annotated pairs from TimeRFT using two external judges, Qwen3-VL-32B~\cite{bai2025qwen3} and GPT-5~\cite{2025gpt5}, neither of which participates in training. We report the verification pass rate that measures the proportion of pairs for which the judge confirms that the query accurately describes the corresponding moment. \begin{wrapfigure}[14]{r}{0.45\columnwidth}
    \centering
    \vspace{-3mm}
    \includegraphics[width=\linewidth]{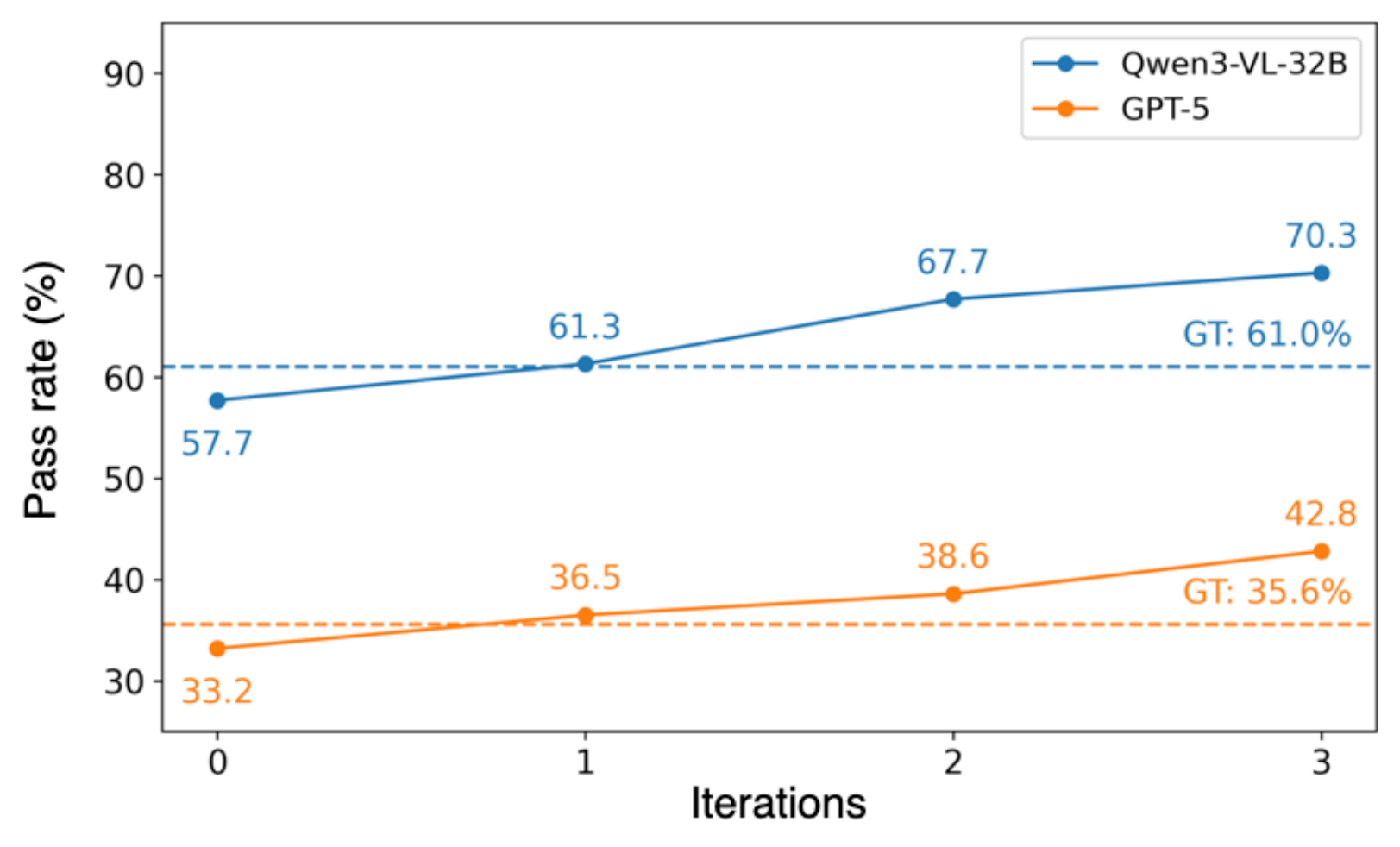}
    \caption{\textbf{Verification pass rate across different iterations and judge models.}}
    \label{fig:pass_rate}
\end{wrapfigure}Although generated pairs by EvoGround and ground-truth annotations share the same source videos, they describe different events and temporal segments. Thus, this analysis is not a direct comparison of data quality; a higher verification pass rate should not be interpreted as evidence of superior data quality. For efficiency, we sample 1,000 and 100 pairs for evaluation by Qwen3-VL-32B and GPT-5, respectively. In Figure~\ref{fig:pass_rate}, pass rates increase monotonically across iterations under both judges. The untrained proposer starts below 60\%, but reaches 70.3\% by iteration 3, suggesting that our proposer and its data evolves across iterations, rather than remaining static or becoming increasingly unfaithful. Since both our proposer and the external judges operate exclusively on sampled video frames, human annotations that rely on non-visual cues, such as audio or speech, may receive lower verification scores despite being correctly annotated.

\subsection{Controls: Does Co-evolution Drive the Gains?}
\label{appendix:sec:controls}
Section~\ref{sec:analysis} reports the headline comparison (Table~\ref{tbl:coevolution}). We
give the settings here. All controls use the same 2.5K TimeRFT videos, the same
Qwen2.5-VL-7B backbone and GRPO as the solver objective, so that only the data pipeline
varies. \textbf{B0} (random moment + captioning) samples moments uniformly from each video
and prompts the frozen base model to caption each sampled clip, yielding pseudo
query--moment pairs with no proposer at all. \textbf{B1} (one-shot pseudo-label) uses the
frozen base model as the proposer, generating pairs once with a tailored prompt and no
solver feedback, after which the solver is trained via GRPO. We keep GRPO rather than SFT
so that the objective is identical across rows.
\textbf{B2} (frozen proposer) bootstraps the proposer for one iteration with the format
reward only and then freezes it, training the solver on its fixed output for the full
budget. These correspond to a basic caption-and-pseudo-label pipeline, one-shot pseudo
query--moment generation, and a frozen proposer without iterative feedback, respectively.
Matching is on solver update steps, so every row receives the same number of gradient steps
on the solver. The only difference is whether the generator is updated.

\subsection{Isolating the Reward Design from the Self-Play Scaffold}
\label{appendix:subsec:rzero}
To separate our contribution from the challenger--solver scaffold itself, we hold the
scaffold fixed and swap in R-Zero's~\cite{huang2025rzero} reward recipe. Since R-Zero's
released models are text-only reasoning LLMs, direct initialization from R-Zero is not
applicable. We therefore train an R-Zero-style proposer with the same GDPO machinery, the
same backbone and the same TimeRFT videos, but with R-Zero's reward composition: format plus
an uncertainty-band solvability term $1 - 2\cdot|\hat{p} - 0.5|$, where $\hat{p}$ is the
solve rate over the completion's proposed windows, each binarized at $\mathrm{tIoU} \geq 0.5$,
and with no consistency term. Both proposers are initialized from the iteration-1 checkpoint
and trained for 150 optimizer steps, single seed, single GPU, and we score their generated
pairs with the external Qwen3-VL-32B judge of Section~\ref{sec:analysis}.

\begin{table*}[h]
    \centering
    \caption{\textbf{Reward design vs.\ self-play scaffold.} Both rows use the same scaffold,
    backbone and corpus, and differ only in the proposer's reward composition. Short-budget
    comparison (150 optimizer steps), not fully trained models.}
    \label{appendix:tbl:rzero}
    \resizebox{\linewidth}{!}{
    \begin{tabular}{l l c}
    \toprule
    \textbf{Method} & \textbf{Reward composition} & \textbf{Qwen3-VL-32B pass \% (conf.)} \\
    \midrule
    R-Zero-style~\cite{huang2025rzero} & format + solver-uncertainty band, no faithfulness & 58.2 (0.51) \\
    \bf \method~reward & format + consistency + solver feedback & \bf 63.4 (0.56) \\
    \bottomrule
    \end{tabular}
    }
\end{table*}

The R-Zero-style reward produces less faithful pairs at a matched budget, quantifying the
contribution of the VTG-specific reward design beyond the self-play scaffold.

\subsection{Self-Evolved Solver as Initialization for Supervised Fine-Tuning}
\label{appendix:subsec:sft_init}
A natural question is whether the self-evolution loop yields a better starting point for
supervised adaptation than the raw backbone. We fine-tune on the 12K Charades-STA training
labels starting from our self-evolved solver, keeping GRPO as the learning objective so that
only the initialization differs, and compare against identical fine-tuning from the raw
backbone.

\begin{table*}[h]
    \centering
    \caption{\textbf{Supervised fine-tuning from the self-evolved solver} on Charades-STA
    (R1@0.3 / R1@0.5 / R1@0.7 / mIoU). Only the initialization differs between the first and
    last rows.}
    \label{appendix:tbl:sft_init}
    \resizebox{0.7\linewidth}{!}{
    \begin{tabular}{l c}
    \toprule
    \textbf{Initialization} & \textbf{Charades-STA} \\
    \midrule
    Qwen2.5-VL + FT & 76.1 / 42.9 / 26.2 / 43.6 \\
    Time-R1~\cite{wang2025timer1} (reported) & \bf 82.8 / 72.2 / \bf 50.1 / 60.9 \\
    \bf \method~+ FT & 80.9 / \bf 72.9 / 48.9 / \bf 61.3 \\
    \bottomrule
    \end{tabular}
    }
\end{table*}

Initializing from the self-evolved solver substantially outperforms initializing from the raw
backbone under an otherwise identical fine-tuning setup ($+17.7$ mIoU), indicating that the
loop provides a stronger starting point for supervised adaptation. The resulting model
attains the best mIoU and R1@0.5 overall while forgoing Time-R1's cold-start SFT and dynamic
hard-sample mining.

\subsection{Efficiency Comparison with Annotation-Based Competitors}
\label{appendix:subsec:efficiency}

\begin{table*}[t]
    \centering
    \caption{\textbf{Training cost and supervision.} \method~trains with substantially less
    compute than ED-VTG and fewer GPUs than Time-R1 while requiring no labeled pairs. The
    annotation cost is not shifted to computation.}
    \label{appendix:tbl:efficiency}
    \resizebox{\linewidth}{!}{
    \begin{tabular}{l l l l}
    \toprule
     & \bf \method & \bf Time-R1~\cite{wang2025timer1} & \bf ED-VTG~\cite{pramanick2025enrich} \\
    \midrule
    Params & 7B & 7B & 7B \\
    External models & SigLIP-2 (0.4B) & - & LLaVA-OneVision (72B) \\
    Annotations & \bf 0 labeled pairs & $\sim$80K labeled pairs & $\sim$136K labeled pairs \\
    Pipeline stages & no cold-start SFT, no CoT curation, & cold-start SFT + dynamic & SFT \\
                    & no hard-sample mining & hard-sample mining & \\
    Training cost & $\sim$72\,h wall-clock (3 iterations $\times$ $\sim$24\,h) & 8$\times$A100 GPUs, & $\sim$2 days on 16 nodes of \\
                        & $\approx$288 GPU-hours on 4$\times$A100-80GB & training time not reported & 8$\times$V100-32GB $\approx$ 6{,}144 GPU-hours \\
    Peak memory & $\sim$60\,GB/GPU (proposer), $\sim$30\,GB/GPU (solver) & - & - \\
    \bottomrule
    \end{tabular}
    }
\end{table*}

Per-iteration breakdown: proposer $\sim$14\,h, solver $\sim$10\,h, pair generation $\sim$1\,h.
Because frames are sampled at 2 FPS and adaptively resized to a fixed pixel budget,
per-video inference cost is roughly constant with video length, and cost scales linearly with
corpus size (cf.\ Table~\ref{tbl:scaling}, 2.5K $\rightarrow$ 7.5K).

\subsection{Comparison with Pseudo-Label and Zero-Shot Grounding Pipelines}
\label{appendix:subsec:zeroshot_prior}

\begin{table*}[h]
    \centering
    \caption{
    \textbf{Comparison with annotation-free grounding pipelines} under the zero-shot
    protocol (R1@0.3 / R1@0.5 / R1@0.7 / mIoU). Numbers for ReCorrect and BTDP are taken from
    their respective papers. Unlike these methods, which train specialized grounding heads
    that output temporal intervals only, \method~yields a general-purpose Video-LLM that also
    improves broader video understanding such as dense captioning
    (Table~\ref{tbl:dvc}).
    }
    \label{appendix:tbl:zeroshot_prior}
    \begin{tabular}{l c c}
    \toprule
    \textbf{Method} & \textbf{Charades-STA} & \textbf{ActivityNet-Captions} \\
    \midrule
    ReCorrect~\cite{bao2025vidgroup} & 66.5 / 51.2 / 28.5 / 45.6 & 54.7 / 33.4 / 15.2 / 36.0 \\
    BTDP~\cite{deng2025btdp}        & 58.3 / 40.0 / 20.9 / 39.0 & 50.6 / 30.5 / 17.5 / 36.5 \\
    \bf \method                      & \bf 77.2 / 60.5 / 35.5 / 53.1 & \bf 62.9 / 43.6 / 25.0 / 43.9 \\
    \bottomrule
    \end{tabular}
\end{table*}

ReCorrect pretrains on 50K videos with 200K pseudo labels from a fixed generation
pipeline, then filters mismatched pairs and corrects boundaries from a memory of its own
predictions during pretraining. The generating pipeline itself is never updated. BTDP likewise
relies on a hand-designed pipeline with no trained components. In contrast, \method~mutually evolves
both agents on 2.5K videos with no pretraining stage. We exclude R-17pV's reference [A],
graph-convolutional temporal action localization~\cite{zeng2019gcn}, from this table because
the protocols do not align: it performs closed-set action localization evaluated by mAP
rather than language-conditioned grounding evaluated by R1@IoU.

\subsection{Captioning Evaluation Protocol}
\label{appendix:subsec:caption_protocol}
To verify that the captioning results of Table~\ref{tbl:temporalbench} reflect the model
rather than the evaluation pipeline, we evaluate in both directions: (1) the base model is
prompted with our proposer's prompt and its per-segment outputs are concatenated through the
identical pipeline used for \method, and (2) both models are evaluated under the single-pass
protocol of the original TemporalBench release, where each model generates one description of
the video.

\begin{table*}[h]
    \centering
    \caption{\textbf{TemporalBench under both protocols} (Similarity / CIDEr / ROUGE). The gap
    between protocols is negligible and Similarity is near-saturated. The more discriminative
    metrics favour \method~under both.}
    \label{appendix:tbl:caption_protocol}
    \resizebox{0.7\linewidth}{!}{
    \begin{tabular}{l cc}
    \toprule
    \textbf{Method} & \textbf{Concatenation} & \textbf{Single-pass} \\
    \midrule
    Qwen2.5-VL-7B~\cite{bai2025qwen2} & 52.4 / 9.8 / 19.3 & \bf 53.8 / 9.9 / 19.5 \\
    \bf \method & \bf 53.8 / 11.4 / 20.5 & 53.7 / \bf 11.1 / \bf 19.8 \\
    \bottomrule
    \end{tabular}
    }
\end{table*}

\subsection{Preference Study Against Human Annotations on Matched Moments}
\label{appendix:subsec:preference}
The pass-rate audit in Section~\ref{sec:analysis} scores generated and human-annotated pairs
independently. To compare them on the \emph{same} temporal content, we additionally match each
generated pair to its most similar ground-truth pair on the same video by
$\mathrm{IoU} \geq 0.7$, keeping only matches whose moment lengths differ by less than 30\%.
This yields at least 300 matched pairs per iteration. We judge a seeded random sample of 100
per iteration. The judge (Qwen3-VL-32B-Instruct) sees the union clip of the two moments with 8
uniformly sampled frames and is asked which query better describes the clip. To control for
position bias, each pair is judged twice with the two options swapped. A win or loss is counted
only when both orders agree, and disagreements count as ties.

\begin{table*}[h]
    \centering
    \caption{\textbf{Preference study on temporally matched pairs.} Strict counting: a win
    requires both presentation orders to agree, and ties count against \method. The judge
    selected option (a) 45.5--51\% of the time, indicating no appreciable position bias.}
    \label{appendix:tbl:preference}
    \resizebox{0.8\linewidth}{!}{
    \begin{tabular}{l cccc}
    \toprule
    \textbf{Comparison} & \textbf{Win \% (strict)} & \textbf{Tie \%} & \textbf{Loss \%} & \textbf{Win \% (decisive only)} \\
    \midrule
    Iteration 1 vs.\ GT & 58.0 & 14.0 & 28.0 & 67.4 \\
    Iteration 2 vs.\ GT & 62.0 & 19.0 & 19.0 & 76.5 \\
    Iteration 3 vs.\ GT & 66.0 & 12.0 & 22.0 & 75.0 \\
    \bottomrule
    \end{tabular}
    }
\end{table*}

We do not read these results as evidence that generated queries are better than human
annotations. Automatic judges are known to favor longer, more detailed descriptions, which
disadvantages ground-truth queries that rely on audio or full-video context, and the tie rates
(12--19\%) themselves support ``comparable'' over ``strictly better.'' The consistent
$\geq$50\% preference indicates only that, for nearly identical temporal moments, our
generated queries are at least comparable to human-written annotations in visual faithfulness
and specificity.

\subsection{Pretrained Components and Potential Benchmark Exposure}
\label{appendix:subsec:exposure}
For transparency we list every pretrained component: Qwen2.5-VL-7B (plus Qwen2.5-VL-3B,
Qwen3-VL-8B and InternVL3.5-4B in the backbone ablation), SigLIP-2 (0.4B) for the consistency
reward, and the 2.5K-video TimeRFT corpus. Qwen's pretraining corpus is not fully disclosed,
so prior exposure to ActivityNet or Charades videos and captions cannot be categorically
excluded. Two observations bound how far this can affect our conclusions. First, the zero-shot
backbone performs far worse than \method~(28.1 $\rightarrow$ 42.1 R1@0.3 on TVGBench,
Table~\ref{tbl:backbone_ablation}, and the Charades-STA rows of
Table~\ref{tbl:charades&activity}), so benchmark memorization cannot explain the gains.
Second, every RL baseline we compare against (Time-R1, VideoChat-R1/R1.5) shares the same
backbone, so any leakage advantage is common-mode across the compared methods.

\subsection{Controlling the Moment-Length Distribution}
\label{appendix:subsec:moment_length}
Several components influence the length distribution of the generated moments: (i)
$\tau_{\text{length}}$, the minimum segment length in the validity constraint, (ii) the
duration constraints stated in the proposer prompt (Figure~\ref{appendix:fig:prompt}), and
(iii) the frame-budget clip $T_{\min} = 8$, $T_{\max} = 32$. Empirically, the
consistency-reward weight also tightens the distribution (27.0 $\pm$ 37.4 $\rightarrow$
19.8 $\pm$ 24.6, Figure~\ref{fig:histogram}), so the framework can be biased toward shorter or longer target moments through these knobs.
\endgroup

\newpage
\clearpage
\section*{NeurIPS Paper Checklist}

\begin{enumerate}

\item {\bf Claims}
    \item[] Question: Do the main claims made in the abstract and introduction accurately reflect the paper's contributions and scope?
    \item[] Answer: \answerYes{}
    \item[] Justification: Abstract and introduction clearly include the main claims.
    \item[] Guidelines:
    \begin{itemize}
        \item The answer \answerNA{} means that the abstract and introduction do not include the claims made in the paper.
        \item The abstract and/or introduction should clearly state the claims made, including the contributions made in the paper and important assumptions and limitations. A \answerNo{} or \answerNA{} answer to this question will not be perceived well by the reviewers. 
        \item The claims made should match theoretical and experimental results, and reflect how much the results can be expected to generalize to other settings. 
        \item It is fine to include aspirational goals as motivation as long as it is clear that these goals are not attained by the paper. 
    \end{itemize}

\item {\bf Limitations}
    \item[] Question: Does the paper discuss the limitations of the work performed by the authors?
    \item[] Answer: \answerYes{} 
    \item[] Justification: The paper discusses the limitations of the work in Section~\ref{sec:conclusion}.
    \item[] Guidelines:
    \begin{itemize}
        \item The answer \answerNA{} means that the paper has no limitation while the answer \answerNo{} means that the paper has limitations, but those are not discussed in the paper. 
        \item The authors are encouraged to create a separate ``Limitations'' section in their paper.
        \item The paper should point out any strong assumptions and how robust the results are to violations of these assumptions (e.g., independence assumptions, noiseless settings, model well-specification, asymptotic approximations only holding locally). The authors should reflect on how these assumptions might be violated in practice and what the implications would be.
        \item The authors should reflect on the scope of the claims made, e.g., if the approach was only tested on a few datasets or with a few runs. In general, empirical results often depend on implicit assumptions, which should be articulated.
        \item The authors should reflect on the factors that influence the performance of the approach. For example, a facial recognition algorithm may perform poorly when image resolution is low or images are taken in low lighting. Or a speech-to-text system might not be used reliably to provide closed captions for online lectures because it fails to handle technical jargon.
        \item The authors should discuss the computational efficiency of the proposed algorithms and how they scale with dataset size.
        \item If applicable, the authors should discuss possible limitations of their approach to address problems of privacy and fairness.
        \item While the authors might fear that complete honesty about limitations might be used by reviewers as grounds for rejection, a worse outcome might be that reviewers discover limitations that aren't acknowledged in the paper. The authors should use their best judgment and recognize that individual actions in favor of transparency play an important role in developing norms that preserve the integrity of the community. Reviewers will be specifically instructed to not penalize honesty concerning limitations.
    \end{itemize}

\item {\bf Theory assumptions and proofs}
    \item[] Question: For each theoretical result, does the paper provide the full set of assumptions and a complete (and correct) proof?
    \item[] Answer: \answerNA{} 
    \item[] Justification: The content does not include theoretical results, requiring proof.
    \item[] Guidelines:
    \begin{itemize}
        \item The answer \answerNA{} means that the paper does not include theoretical results. 
        \item All the theorems, formulas, and proofs in the paper should be numbered and cross-referenced.
        \item All assumptions should be clearly stated or referenced in the statement of any theorems.
        \item The proofs can either appear in the main paper or the supplemental material, but if they appear in the supplemental material, the authors are encouraged to provide a short proof sketch to provide intuition. 
        \item Inversely, any informal proof provided in the core of the paper should be complemented by formal proofs provided in appendix or supplemental material.
        \item Theorems and Lemmas that the proof relies upon should be properly referenced. 
    \end{itemize}

    \item {\bf Experimental result reproducibility}
    \item[] Question: Does the paper fully disclose all the information needed to reproduce the main experimental results of the paper to the extent that it affects the main claims and/or conclusions of the paper (regardless of whether the code and data are provided or not)?
    \item[] Answer: \answerYes{} 
    \item[] Justification: The paper provides detailed experiment settings for reproducibility.
    \item[] Guidelines:
    \begin{itemize}
        \item The answer \answerNA{} means that the paper does not include experiments.
        \item If the paper includes experiments, a \answerNo{} answer to this question will not be perceived well by the reviewers: Making the paper reproducible is important, regardless of whether the code and data are provided or not.
        \item If the contribution is a dataset and\slash or model, the authors should describe the steps taken to make their results reproducible or verifiable. 
        \item Depending on the contribution, reproducibility can be accomplished in various ways. For example, if the contribution is a novel architecture, describing the architecture fully might suffice, or if the contribution is a specific model and empirical evaluation, it may be necessary to either make it possible for others to replicate the model with the same dataset, or provide access to the model. In general. releasing code and data is often one good way to accomplish this, but reproducibility can also be provided via detailed instructions for how to replicate the results, access to a hosted model (e.g., in the case of a large language model), releasing of a model checkpoint, or other means that are appropriate to the research performed.
        \item While NeurIPS does not require releasing code, the conference does require all submissions to provide some reasonable avenue for reproducibility, which may depend on the nature of the contribution. For example
        \begin{enumerate}
            \item If the contribution is primarily a new algorithm, the paper should make it clear how to reproduce that algorithm.
            \item If the contribution is primarily a new model architecture, the paper should describe the architecture clearly and fully.
            \item If the contribution is a new model (e.g., a large language model), then there should either be a way to access this model for reproducing the results or a way to reproduce the model (e.g., with an open-source dataset or instructions for how to construct the dataset).
            \item We recognize that reproducibility may be tricky in some cases, in which case authors are welcome to describe the particular way they provide for reproducibility. In the case of closed-source models, it may be that access to the model is limited in some way (e.g., to registered users), but it should be possible for other researchers to have some path to reproducing or verifying the results.
        \end{enumerate}
    \end{itemize}

\item {\bf Open access to data and code}
    \item[] Question: Does the paper provide open access to the data and code, with sufficient instructions to faithfully reproduce the main experimental results, as described in supplemental material?
    \item[] Answer: \answerYes{} 
    \item[] Justification: We will make all resources publicly available.
    \item[] Guidelines:
    \begin{itemize}
        \item The answer \answerNA{} means that paper does not include experiments requiring code.
        \item Please see the NeurIPS code and data submission guidelines (\url{https://neurips.cc/public/guides/CodeSubmissionPolicy}) for more details.
        \item While we encourage the release of code and data, we understand that this might not be possible, so \answerNo{} is an acceptable answer. Papers cannot be rejected simply for not including code, unless this is central to the contribution (e.g., for a new open-source benchmark).
        \item The instructions should contain the exact command and environment needed to run to reproduce the results. See the NeurIPS code and data submission guidelines (\url{https://neurips.cc/public/guides/CodeSubmissionPolicy}) for more details.
        \item The authors should provide instructions on data access and preparation, including how to access the raw data, preprocessed data, intermediate data, and generated data, etc.
        \item The authors should provide scripts to reproduce all experimental results for the new proposed method and baselines. If only a subset of experiments are reproducible, they should state which ones are omitted from the script and why.
        \item At submission time, to preserve anonymity, the authors should release anonymized versions (if applicable).
        \item Providing as much information as possible in supplemental material (appended to the paper) is recommended, but including URLs to data and code is permitted.
    \end{itemize}

\item {\bf Experimental setting/details}
    \item[] Question: Does the paper specify all the training and test details (e.g., data splits, hyperparameters, how they were chosen, type of optimizer) necessary to understand the results?
    \item[] Answer: \answerYes{} 
    \item[] Justification: The paper specifies all details for training and testing.
    \item[] Guidelines:
    \begin{itemize}
        \item The answer \answerNA{} means that the paper does not include experiments.
        \item The experimental setting should be presented in the core of the paper to a level of detail that is necessary to appreciate the results and make sense of them.
        \item The full details can be provided either with the code, in appendix, or as supplemental material.
    \end{itemize}

\item {\bf Experiment statistical significance}
    \item[] Question: Does the paper report error bars suitably and correctly defined or other appropriate information about the statistical significance of the experiments?
    \item[] Answer: \answerYes{} 
    \item[] Justification: We report 95\% bootstrap confidence intervals (1,000 resamples with replacement of the test set) for our model in Tables~\ref{tbl:charades&activity} and~\ref{tbl:tvgbench&rextime&etbench}, except for ReXTime, whose ground truth is held on the official test server. Methodology is described in Section~\ref{subsec:main_results} (Evaluation Metrics).
    \item[] Guidelines:
    \begin{itemize}
        \item The answer \answerNA{} means that the paper does not include experiments.
        \item The authors should answer \answerYes{} if the results are accompanied by error bars, confidence intervals, or statistical significance tests, at least for the experiments that support the main claims of the paper.
        \item The factors of variability that the error bars are capturing should be clearly stated (for example, train/test split, initialization, random drawing of some parameter, or overall run with given experimental conditions).
        \item The method for calculating the error bars should be explained (closed form formula, call to a library function, bootstrap, etc.)
        \item The assumptions made should be given (e.g., Normally distributed errors).
        \item It should be clear whether the error bar is the standard deviation or the standard error of the mean.
        \item It is OK to report 1-sigma error bars, but one should state it. The authors should preferably report a 2-sigma error bar than state that they have a 96\% CI, if the hypothesis of Normality of errors is not verified.
        \item For asymmetric distributions, the authors should be careful not to show in tables or figures symmetric error bars that would yield results that are out of range (e.g., negative error rates).
        \item If error bars are reported in tables or plots, the authors should explain in the text how they were calculated and reference the corresponding figures or tables in the text.
    \end{itemize}

\item {\bf Experiments compute resources}
    \item[] Question: For each experiment, does the paper provide sufficient information on the computer resources (type of compute workers, memory, time of execution) needed to reproduce the experiments?
    \item[] Answer: \answerYes{} 
    \item[] Justification: The paper provides such details to reproduce the experiments.
    \item[] Guidelines:
    \begin{itemize}
        \item The answer \answerNA{} means that the paper does not include experiments.
        \item The paper should indicate the type of compute workers CPU or GPU, internal cluster, or cloud provider, including relevant memory and storage.
        \item The paper should provide the amount of compute required for each of the individual experimental runs as well as estimate the total compute. 
        \item The paper should disclose whether the full research project required more compute than the experiments reported in the paper (e.g., preliminary or failed experiments that didn't make it into the paper). 
    \end{itemize}
    
\item {\bf Code of ethics}
    \item[] Question: Does the research conducted in the paper conform, in every respect, with the NeurIPS Code of Ethics \url{https://neurips.cc/public/EthicsGuidelines}?
    \item[] Answer: \answerYes{} 
    \item[] Justification: The research follows the NeurIPS Code of Ethics.
    \item[] Guidelines:
    \begin{itemize}
        \item The answer \answerNA{} means that the authors have not reviewed the NeurIPS Code of Ethics.
        \item If the authors answer \answerNo, they should explain the special circumstances that require a deviation from the Code of Ethics.
        \item The authors should make sure to preserve anonymity (e.g., if there is a special consideration due to laws or regulations in their jurisdiction).
    \end{itemize}

\item {\bf Broader impacts}
    \item[] Question: Does the paper discuss both potential positive societal impacts and negative societal impacts of the work performed?
    \item[] Answer: \answerYes{} 
    \item[] Justification: While \method~is designed for academic video understanding, the ability to automatically retrieve and localize video content without annotations could be misused for surveillance or privacy-invasive monitoring. We acknowledge this risk and note that our model and code will be released under a license prohibiting malicious use.
    \item[] Guidelines:
    \begin{itemize}
        \item The answer \answerNA{} means that there is no societal impact of the work performed.
        \item If the authors answer \answerNA{} or \answerNo, they should explain why their work has no societal impact or why the paper does not address societal impact.
        \item Examples of negative societal impacts include potential malicious or unintended uses (e.g., disinformation, generating fake profiles, surveillance), fairness considerations (e.g., deployment of technologies that could make decisions that unfairly impact specific groups), privacy considerations, and security considerations.
        \item The conference expects that many papers will be foundational research and not tied to particular applications, let alone deployments. However, if there is a direct path to any negative applications, the authors should point it out. For example, it is legitimate to point out that an improvement in the quality of generative models could be used to generate Deepfakes for disinformation. On the other hand, it is not needed to point out that a generic algorithm for optimizing neural networks could enable people to train models that generate Deepfakes faster.
        \item The authors should consider possible harms that could arise when the technology is being used as intended and functioning correctly, harms that could arise when the technology is being used as intended but gives incorrect results, and harms following from (intentional or unintentional) misuse of the technology.
        \item If there are negative societal impacts, the authors could also discuss possible mitigation strategies (e.g., gated release of models, providing defenses in addition to attacks, mechanisms for monitoring misuse, mechanisms to monitor how a system learns from feedback over time, improving the efficiency and accessibility of ML).
    \end{itemize}
    
\item {\bf Safeguards}
    \item[] Question: Does the paper describe safeguards that have been put in place for responsible release of data or models that have a high risk for misuse (e.g., pre-trained language models, image generators, or scraped datasets)?
    \item[] Answer: \answerYes{} 
    \item[] Justification: We build only on publicly released pre-trained models and datasets, and we release the trained \method~model and the generated query--moment pairs under a license prohibiting malicious use (see Question 10), which is the safeguard we rely on for the released assets.
    \item[] Guidelines:
    \begin{itemize}
        \item The answer \answerNA{} means that the paper poses no such risks.
        \item Released models that have a high risk for misuse or dual-use should be released with necessary safeguards to allow for controlled use of the model, for example by requiring that users adhere to usage guidelines or restrictions to access the model or implementing safety filters. 
        \item Datasets that have been scraped from the Internet could pose safety risks. The authors should describe how they avoided releasing unsafe images.
        \item We recognize that providing effective safeguards is challenging, and many papers do not require this, but we encourage authors to take this into account and make a best faith effort.
    \end{itemize}

\item {\bf Licenses for existing assets}
    \item[] Question: Are the creators or original owners of assets (e.g., code, data, models), used in the paper, properly credited and are the license and terms of use explicitly mentioned and properly respected?
    \item[] Answer: \answerYes{} 
    \item[] Justification: The paper properly left all references.
    \item[] Guidelines:
    \begin{itemize}
        \item The answer \answerNA{} means that the paper does not use existing assets.
        \item The authors should cite the original paper that produced the code package or dataset.
        \item The authors should state which version of the asset is used and, if possible, include a URL.
        \item The name of the license (e.g., CC-BY 4.0) should be included for each asset.
        \item For scraped data from a particular source (e.g., website), the copyright and terms of service of that source should be provided.
        \item If assets are released, the license, copyright information, and terms of use in the package should be provided. For popular datasets, \url{paperswithcode.com/datasets} has curated licenses for some datasets. Their licensing guide can help determine the license of a dataset.
        \item For existing datasets that are re-packaged, both the original license and the license of the derived asset (if it has changed) should be provided.
        \item If this information is not available online, the authors are encouraged to reach out to the asset's creators.
    \end{itemize}

\item {\bf New assets}
    \item[] Question: Are new assets introduced in the paper well documented and is the documentation provided alongside the assets?
    \item[] Answer: \answerYes{} 
    \item[] Justification: We release the trained EvoGround model and generated query--moment pairs as new assets. These will be made publicly available with documentation on training details, intended use, and licensing upon acceptance.
    \item[] Guidelines:
    \begin{itemize}
        \item The answer \answerNA{} means that the paper does not release new assets.
        \item Researchers should communicate the details of the dataset\slash code\slash model as part of their submissions via structured templates. This includes details about training, license, limitations, etc. 
        \item The paper should discuss whether and how consent was obtained from people whose asset is used.
        \item At submission time, remember to anonymize your assets (if applicable). You can either create an anonymized URL or include an anonymized zip file.
    \end{itemize}

\item {\bf Crowdsourcing and research with human subjects}
    \item[] Question: For crowdsourcing experiments and research with human subjects, does the paper include the full text of instructions given to participants and screenshots, if applicable, as well as details about compensation (if any)? 
    \item[] Answer: \answerNA{} 
    \item[] Justification: The paper does not conduct crowdsourcing.
    \item[] Guidelines:
    \begin{itemize}
        \item The answer \answerNA{} means that the paper does not involve crowdsourcing nor research with human subjects.
        \item Including this information in the supplemental material is fine, but if the main contribution of the paper involves human subjects, then as much detail as possible should be included in the main paper. 
        \item According to the NeurIPS Code of Ethics, workers involved in data collection, curation, or other labor should be paid at least the minimum wage in the country of the data collector. 
    \end{itemize}

\item {\bf Institutional review board (IRB) approvals or equivalent for research with human subjects}
    \item[] Question: Does the paper describe potential risks incurred by study participants, whether such risks were disclosed to the subjects, and whether Institutional Review Board (IRB) approvals (or an equivalent approval/review based on the requirements of your country or institution) were obtained?
    \item[] Answer: \answerNA{} 
    \item[] Justification: The paper does not conduct crowdsourcing.
    \item[] Guidelines:
    \begin{itemize}
        \item The answer \answerNA{} means that the paper does not involve crowdsourcing nor research with human subjects.
        \item Depending on the country in which research is conducted, IRB approval (or equivalent) may be required for any human subjects research. If you obtained IRB approval, you should clearly state this in the paper. 
        \item We recognize that the procedures for this may vary significantly between institutions and locations, and we expect authors to adhere to the NeurIPS Code of Ethics and the guidelines for their institution. 
        \item For initial submissions, do not include any information that would break anonymity (if applicable), such as the institution conducting the review.
    \end{itemize}

\item {\bf Declaration of LLM usage}
    \item[] Question: Does the paper describe the usage of LLMs if it is an important, original, or non-standard component of the core methods in this research? Note that if the LLM is used only for writing, editing, or formatting purposes and does \emph{not} impact the core methodology, scientific rigor, or originality of the research, declaration is not required.
    \item[] Answer: \answerYes{} 
    \item[] Justification: LLMs are a core component of \method: both the proposer and solver are built on Qwen2.5-VL-7B~\cite{bai2025qwen2}, a large vision-language model, and are optimized via reinforcement learning to perform video temporal grounding without human annotations.
    \item[] Guidelines:
    \begin{itemize}
        \item The answer \answerNA{} means that the core method development in this research does not involve LLMs as any important, original, or non-standard components.
        \item Please refer to our LLM policy in the NeurIPS handbook for what should or should not be described.
    \end{itemize}

\end{enumerate}
\newpage

\end{document}